%% file: main_journal.tex
\documentclass[10pt,a4paper]{article}

\usepackage[a4paper,top=2.5cm,bottom=2.5cm,left=2.5cm,right=2.5cm,headsep=0.8cm]{geometry}
\usepackage[T1]{fontenc}
\usepackage[utf8]{inputenc}
\usepackage{newpxtext,newpxmath}   
\usepackage{inconsolata}
\usepackage{microtype}
\usepackage{graphicx}
\usepackage{booktabs}
\usepackage{array}
\usepackage{tikz}
\usepackage{placeins}
\usepackage{amsmath}
\usepackage{multirow}
\usepackage{xcolor}
\usepackage{adjustbox}
\usepackage{makecell}
\usepackage{url}

\usepackage[numbers,square,sort&compress]{natbib}
\usepackage{titlesec}
\titleformat{\section}{\normalfont\bfseries}{\thesection.}{0.5em}{}
\titleformat{\subsection}{\normalfont\itshape}{\thesubsection.}{0.5em}{}
\titleformat{\paragraph}[runin]{\normalfont\bfseries}{}{0pt}{}
\titlespacing*{\section}{0pt}{12pt}{6pt}
\titlespacing*{\subsection}{0pt}{10pt}{4pt}
\titlespacing*{\paragraph}{0pt}{6pt}{0.6em}
\usepackage[font=small,labelfont=bf,labelsep=period,justification=justified,singlelinecheck=false]{caption}
\usepackage{floatrow}
\usepackage{enumitem}
\setlist{topsep=3pt,itemsep=1pt,parsep=0pt}

\usepackage{fancyhdr}
\usepackage{lastpage}
\fancypagestyle{first}{\fancyhf{}\fancyfoot[R]{\small\thepage{} of \pageref{LastPage}}}

\usepackage[colorlinks=true,linkcolor=blue!50!black,citecolor=blue!50!black,urlcolor=blue!50!black]{hyperref}

\newcommand{\paperkeywords}{decision models; zero-shot classification; emotion classification; multilingual; large language models; fine-tuning; statistical significance testing; equivalence testing}
\renewenvironment{abstract}
  {\par\vspace{4pt}\noindent\small\textbf{Abstract:}~\ignorespaces}
  {\par\vspace{6pt}\noindent\textbf{Keywords:}~\paperkeywords\par\vspace{2pt}\noindent\rule{\textwidth}{0.4pt}\par\vspace{8pt}\normalsize}

\newcommand{\backsection}[1]{\par\medskip\noindent\textbf{#1}~}
\newcommand{\paperbibliography}{%
  {\small
  \backsection{Data Availability Statement:} The per-text predictions of every system and the statistical scripts will be released in a public repository upon publication of this article.
  \backsection{Acknowledgments:} The authors used Claude Sonnet 5.5 (Anthropic) to translate the instructions used in the complementary runs (Section~\ref{sec:complementary} and Appendix~\ref{app:prompts}); the translations were checked by one of the authors. The authors take full responsibility for the content of this publication.

  \par}}

\begin{document}
\thispagestyle{first}

\noindent{\LARGE\bfseries\hyphenpenalty=10000\exhyphenpenalty=10000 Decision Models for Multilingual Emotion Classification:\\ A Comparative Study Against LLMs and Encoders\par}
\vspace{10pt}
\noindent{\large Alessandro Piccolo\textsuperscript{\dag}, Francesco De Nicolò\textsuperscript{\dag}, Francesco Manco and Eufemia Lella\textsuperscript{*}\par}
\vspace{6pt}
{\small
\noindent Sirio -- Research \& Innovation, Sidea Group, Bari, Italy\par
\noindent\textsuperscript{\dag}\,These authors contributed equally to this work.\par
\noindent\textsuperscript{*}\,Correspondence: \texttt{eufemia.lella@sideagroup.com}\par}

\input{body}

\clearpage
\bibliography{references}

\end{document}

%% file: body.tex

\providecommand{\statref}{\ifdefined\mdpiversion the Supplementary Materials, Statistical details\else Appendix~\ref{app:statdetails}\fi}
\ifdefined\mdpiversion\else
\begin{abstract}
Decision models receive a question with a closed list of answers and return a probability for each answer instead of free text. They promise low latency and robustness to prompt wording, but no study has compared several of them with fine-tuned encoders across languages. We ask whether systems used without training on labelled data (zero-shot) can replace fine-tuning, and what the decision format adds over a general-purpose large language model (LLM). We compare three decision models (Jev, OpenAI Decisions, Strands Decider 2B) and two LLMs (GPT-6 Luna, Qwen3.5-2B), given Italian instructions, with two fine-tuned multilingual encoders (tabularisai, XLM-EMO) on four emotions (joy, sadness, anger, fear) in 10 social-media datasets in five languages, with paired difference and equivalence tests. The three best zero-shot systems (two decision models and one LLM) score on average about one macro-F1 point below an encoder fine-tuned on the target dataset; our test sets cannot exclude a gap of 4--5 points, and on single datasets either approach wins by up to 22 points. They beat encoders fine-tuned on another dataset by 8--15 points, though against XLM-EMO this may not extend to new datasets, and instructions in each text's language change their results by under 1 point. Exploratorily, they beat the strongest public emotion model used as is by about 6 points. OpenAI's decision endpoint matches GPT-6 Luna, the LLM it runs on, at a fifth of its median latency, and a 2-billion-parameter decision model beats a same-size general LLM by 10 points, owing to its training, not the decision format. The best zero-shot systems, the LLM included, have an SD of about 1 point across prompts. Without labelled data they are a strong default, and decision models add lower latency and, unlike generative LLM use, per-option probabilities. Fine-tuning won by a wide margin only on the largest dataset, GoEmotions.
\end{abstract}
\fi

\section{Introduction}
\label{sec:intro}

Emotion recognition from short texts is a well-established task in natural language processing, with applications ranging from public opinion analysis and customer service support to the detection of mental-health signals in online communities. Research in this area has further been supported by the availability of publicly accessible datasets across multiple languages. A common approach consists of fine-tuning pre-trained multilingual encoder models, such as XLM-R \citep{conneau-etal-2020-unsupervised}, on labelled data from the target domain. An example is XLM-EMO \citep{bianchi-etal-2022-xlm}, developed to recognise four basic emotions across 19 languages. While this approach can achieve strong performance when labelled data are available, collecting and annotating data for each new domain, platform or language is costly. Moreover, models trained on a specific dataset often exhibit limited generalisation to other datasets \citep{bostan-klinger-2018-analysis}.

When labelled data for a target domain are unavailable, several alternative approaches can be considered: reusing a model fine-tuned on a different dataset, directly applying a publicly available emotion model, or adopting a \emph{zero-shot} approach, which requires no task-specific training. Two categories of zero-shot approaches are available. The first comprises general-purpose large language models (LLMs), which can be prompted to assign a label to each input text. Although LLMs achieve performance comparable to that of fine-tuned models on simple sentiment analysis tasks, they tend to underperform on finer-grained affective tasks \citep{zhang-etal-2024-sentiment}. Furthermore, their predictions may be sensitive to the wording of the prompt and to the order of the candidate labels \citep{sclar2024quantifying,zheng2024large,pezeshkpour-hruschka-2024-large}. The second category, several implementations of which were released between September and October 2026, is represented by \emph{decision models}: systems designed to process a question together with a predefined set of candidate options and return a probability distribution over these options rather than generating free-form text \citep{typesafe-jev,openai-decisions,strands-decider}.

By design, decision models are constrained to a predefined set of options, and their output scores can be directly interpreted as confidence values. Their providers also claim low inference latency and robustness to variations in prompt formulation. This class of models is defined by its interface rather than by a specific mechanism. For closed-source systems, the internal inference procedure is not publicly documented and may, for instance, merely involve extracting the probabilities assigned by an LLM to the candidate labels through a dedicated interface. We therefore evaluate these systems as black boxes, exactly as offered. To date, the only independent study, Emo-Jev \citep{emojev}, evaluates a single decision model in single-run experiments, without reporting confidence intervals or conducting statistical significance tests (Section~\ref{sec:related}). To the best of our knowledge, no previous study has systematically compared multiple decision models with fine-tuned encoders across languages while accounting for statistical uncertainty.

This paper fills that gap with a controlled comparison on a common task. We compare five zero-shot systems, three decision models (Jev, OpenAI Decisions and Strands Decider 2B) and two general LLMs (GPT-6 Luna and Qwen3.5-2B), with two multilingual emotion encoders, XLM-EMO and tabularisai, fine-tuned on each dataset. The task is to assign one of four emotions (joy, sadness, anger or fear) to texts from 10 social-media datasets in five languages. Quality is measured with macro-F1, the mean of the F1 scores of the four classes, and differences are given in points, where one point is 0.01 macro-F1. All zero-shot systems receive the same instructions, written once in Italian (Section~\ref{sec:prompts}); complementary runs test instructions in English and in the language of each text.

The study addresses two main questions and one cross-cutting question. The first is whether zero-shot systems, decision models and general LLMs alike, can replace fine-tuning on labelled data (H1 and H2). The second is whether a decision model is preferable to a general LLM for the same task, in quality and latency (H3). The cross-cutting question is how robust all zero-shot systems are to the wording of the prompt and to the order of the options (H4). We do not vary the amount of labelled data: the first question is answered by comparing the zero-shot systems with encoders fine-tuned on the full training split of the target dataset and on other datasets, and, in an exploratory comparison, with public emotion models used as they are. H1, H3 and H4 use a margin of two points: a smaller difference is treated as negligible in practice, and two systems are called equivalent when the data exclude any difference larger than the margin.
\begin{description}
\item[H1] The best zero-shot systems reach the quality of an encoder fine-tuned \emph{in-domain}, that is, on the training split of the dataset on which it is tested: they are equivalent to it within two points. This is the most demanding comparison for a zero-shot system, because the encoder has seen texts from the same source, annotated with the same guidelines. With test sets of the size available here this equivalence cannot be demonstrated (Section~\ref{sec:stats}), so we also report how large a gap the data can exclude.
\item[H2] (a) A fine-tuned encoder loses quality when applied to a dataset other than its training dataset, and (b) zero-shot systems outperform it in that setting. This is the realistic situation of a practitioner who reuses a model trained elsewhere.
\item[H3] For the same task, a decision model is preferable to a general LLM: at least as good in quality and faster. Two pairs make this test partly controlled: OpenAI Decisions runs on the same model as GPT-6 Luna, and Strands Decider 2B is built on the same base model as Qwen3.5-2B.
\item[H4] The quality of zero-shot systems is insensitive to the wording of the question and to the order of the options, in the sense that it varies by less than the margin of two points; a variation that is detectable but smaller than the margin is consistent with H4.
\end{description}

Small differences between systems evaluated on a handful of datasets are easy to over-interpret \citep{dror-etal-2018-hitchhikers}, and the absence of a significant difference is far from a demonstration of equivalence. Every comparison is therefore paired, that is, computed on exactly the same texts, tests equivalence as well as difference, and is corrected for the number of comparisons. Because a mean over ten datasets can hide large differences between them, we also estimate what to expect on a dataset not in this study, with a random-effects meta-analysis. We say that a result holds \emph{on our datasets} when the mean over our test sets is significant, and that it is \emph{generalisable to new datasets} when the meta-analysis also supports it; we use these two labels with this meaning throughout. The hypotheses and the analysis plan were written after the mean macro-F1 of every system was known but before any comparison was tested, and were not preregistered (Appendix~\ref{app:deviations}).

The three zero-shot systems with the highest mean macro-F1, Jev, OpenAI Decisions and GPT-6 Luna, hereafter \emph{the three leading zero-shot systems}, are on average about one point below in-domain fine-tuning, but the data cannot exclude a gap of 4--5 points, so H1 is inconclusive. They outperform encoders fine-tuned on another dataset, an advantage expected to hold on new datasets against tabularisai but shown only on our datasets against XLM-EMO (H2). OpenAI Decisions matches GPT-6 Luna at one fifth of its median latency, and Strands Decider 2B beats Qwen3.5-2B by ten points thanks to its training, not to the decision format (H3). The three leading systems are robust to the prompt and to instructions in the language of the text, Strands Decider 2B narrowly misses the stability threshold, and Qwen3.5-2B is clearly unstable (H4). Table~\ref{tab:verdicts} summarises all outcomes.

The contributions of this work are threefold. First, we provide the first comparison of several decision models with general LLMs and with fine-tuned encoders, on 10 datasets in five languages and with measures of uncertainty. Second, we test what the decision format adds with three tools: two matched pairs of systems, label scoring applied to the small general LLM, and a latency measurement with one request at a time. Third, our analysis separates ``no detectable difference'' from ``demonstrated equivalence'', and results ``on these datasets'' from results ``on a new dataset''; we release the per-text predictions of all systems and the analysis scripts, so that every number in the paper can be recomputed.

After a review of related work (Section~\ref{sec:related}) and of our methods (Section~\ref{sec:methods}), we report the results hypothesis by hypothesis (Section~\ref{sec:results}), discuss them (Section~\ref{sec:discussion}) and conclude (Section~\ref{sec:conclusions}).

\section{Related Work}
\label{sec:related}

This section reviews the work on which each part of our design builds: zero-shot LLMs compared with fine-tuning (H1), the fine-tuned encoders that serve as references and their transfer across corpora (H2), language models used as classifiers over a closed set of labels, from which decision models descend (H3), sensitivity to the prompt (H4), and the statistical comparison of NLP systems.

Whether prompted LLMs can replace fine-tuned models is still debated, and the evidence for emotions points to a gap. A systematic comparison by \citet{zhang-etal-2024-sentiment} found that LLMs approach fine-tuned small models on simple sentiment classification but fall behind on tasks that require structured or fine-grained affective labels. Evaluating ChatGPT on 25 tasks, emotion recognition among them, \citet{kocon2023chatgpt} found it below the best task-specific models, with larger losses on the harder tasks. \citet{bucher2024fine} report that smaller models fine-tuned on application-specific data outperform zero-shot GPT-4 and Claude on every classification task they consider, emotions included, and \citet{meinert2026llms} find that a zero-shot Llama model stays below a fine-tuned RoBERTa on a multi-dataset emotion benchmark, while few-shot prompting closes the gap on some datasets only. These studies predate decision models, mostly use a single prompt and report no test of equivalence; H1 asks the same question for the strongest zero-shot systems available today, with an explicit margin.

Our fine-tuned references are two multilingual encoders already adapted to emotions. XLM-EMO \citep{bianchi-etal-2022-xlm} fine-tunes XLM-T, a version of XLM-R adapted to Twitter \citep{barbieri-etal-2022-xlm,conneau-etal-2020-unsupervised}, on emotion datasets in 19 languages and is a widely used multilingual model for the four basic emotions considered here. The tabularisai model \citep{borisov-2026-tabularisai} is trained instead on synthetic data generated by LLMs in 23 languages. We fine-tune both on each of our datasets and also evaluate the public checkpoints as they are. Fine-tuned transformers are sensitive to the random seed, which affects weight initialisation and data order \citep{dodge2020fine,mosbach2021stability}, so results should be averaged over several seeds, which motivates our three seeds per configuration. Fine-tuning such models further on a single dataset improves in-domain accuracy but can distort the features learned in pre-training and degrade performance out of distribution \citep{kumar2022finetuning}. Emotion corpora also differ in their source (tweets, forum posts, comments), in the annotation scheme and guidelines, and in the distribution of labels: \citet{bostan-klinger-2018-analysis} unified many English emotion corpora and showed that classifiers trained on one corpus transfer poorly to the others. Cross-corpus evaluation, in which a model trained on one dataset is tested on the others, is the protocol adopted by \citet{barbieri-etal-2022-xlm}, which we follow for the fine-tuned encoders.

Decision models belong to a longer line of work that uses language models as classifiers over a closed set of labels. Before instruction-following LLMs, zero-shot classification was often cast as natural language inference, that is, deciding whether the text entails a hypothesis such as ``This text expresses joy'', with one hypothesis per label \citep{yin-etal-2019-benchmarking}; for emotions, \citet{plaza-del-arco-etal-2022-natural} showed that the choice of the hypothesis changes the results considerably and that no single formulation works best across corpora. A generative model can also classify by \emph{label scoring}, that is, by comparing the probabilities it assigns to the words of each label \citep{brown2020language}; verbalizers map each class to such words \citep{schick-schutze-2021-exploiting}, and the resulting label probabilities are biased towards some labels and positions unless calibrated \citep{zhao2021calibrate}. Constrained decoding, finally, restricts generation to a predefined set of outputs \citep{willard2023efficient}. A decision model offers the outcome of these techniques, a probability for each option, as a ready-made format. Strands Decider 2B, in particular, is trained in part by knowledge distillation, in which a smaller model learns to reproduce the outputs of a larger one \citep{hinton2015distilling}. 

The only previous evaluation of a decision model is Emo-Jev \citep{emojev}, which compares Jev with five LLMs on seven English and one Chinese dataset for sentiment, emotion, sarcasm and humour, and also proposes training-free ways of combining several Jev judgements. Plain Jev reaches an average macro-F1 of 0.629 against 0.673 for the strongest LLM, with lower latency, in single runs without measures of uncertainty. Whether the decision format costs quality, as that result suggests, or only saves latency is the question of H3. Label scoring gives a general LLM the output of a decision model without any training, and thus separates the effect of the decision format from that of training (Sections~\ref{sec:complementary} and~\ref{sec:res-h3}); we return to the difference between the two studies in Section~\ref{sec:disc-when}.

The performance of LLMs can change substantially when a prompt is reformulated in semantically equivalent ways or merely reformatted \citep{sclar2024quantifying}. As a consequence, a comparison between models based on a single prompt can be misleading, and \citet{mizrahi-etal-2024-state} recommend evaluating each model over several formulations. In multiple-choice settings, models also show systematic preferences for particular positions or option labels \citep{zhao2021calibrate,zheng2024large,pezeshkpour-hruschka-2024-large}. The language of the instructions matters as well: task descriptions in English often gave ChatGPT better results than descriptions in the language of the text \citep{lai-etal-2023-chatgpt}, and machine-translated prompt templates performed much worse than English ones for an earlier GPT-3.5 model \citep{ahuja-etal-2023-mega}. Choosing the prompt on labelled data, on the other hand, violates the zero-shot assumption \citep{perez2021true}. In line with these findings, we evaluate every zero-shot system over several question variants with balanced option orders (Section~\ref{sec:prompts}), and, since our instructions are in Italian for texts in five languages, we also test them in English and in the language of each text.

Significance testing in NLP is commonly based on paired resampling tests on the same test items \citep{berg-kirkpatrick-etal-2012-empirical}, and \citet{dror-etal-2018-hitchhikers} recommend choosing the test according to the metric and the dependence structure of the data and correcting for multiple comparisons. When systems are compared over several datasets, \citet{demsar2006statistical} recommends non-parametric tests on the per-dataset results, such as the Wilcoxon signed-rank test, and \citet{benavoli2016should} warn against post hoc procedures based on mean ranks, which is why we compare pairs of systems directly instead of ranking them. Two aspects are less common in NLP evaluations and central to our questions, and we use both (Section~\ref{sec:stats}): testing equivalence rather than only difference \citep{schuirmann1987comparison,lakens2017equivalence}, and treating datasets as a sample from a population, as is done in meta-analysis \citep{higgins2009reevaluation}.

\section{Materials and Methods}
\label{sec:methods}

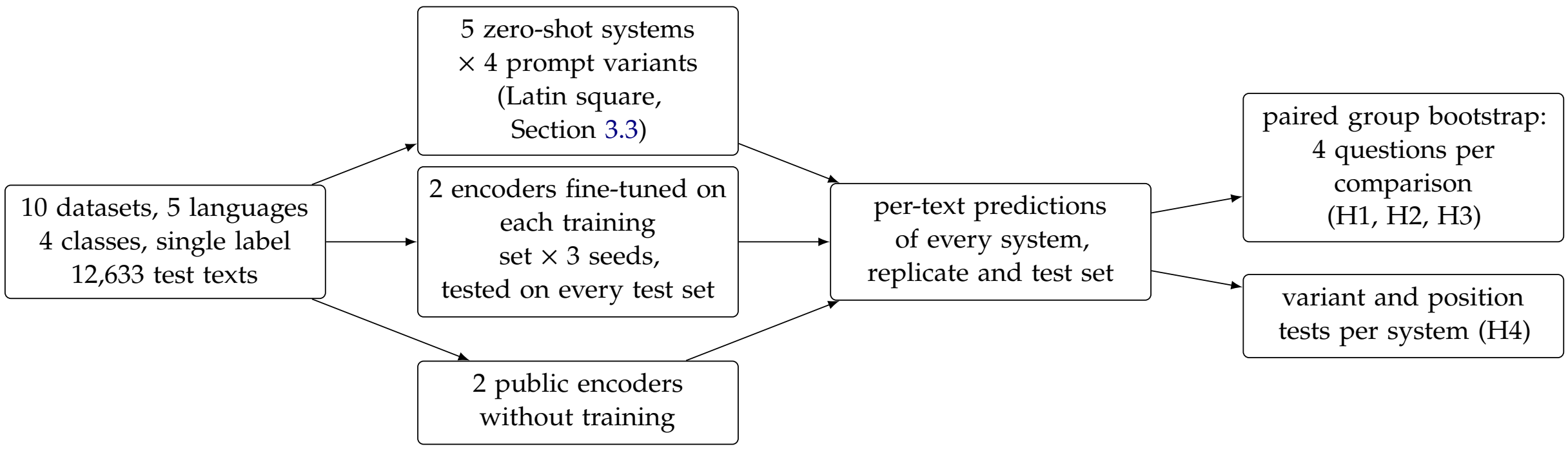
\begin{figure*}[!htb]
\centering
\begin{adjustbox}{max width=\linewidth}
\begin{tikzpicture}[font=\small, box/.style={draw, rounded corners=2pt, align=center, inner sep=4pt, text width=3.6cm}, >=latex]
\node[box] (data) at (0,0) {10 datasets, 5 languages\\4 classes, single label\\12{,}633 test texts};
\node[box] (zs) at (5,1.95) {5 zero-shot systems\\$\times$ 4 prompt variants\\(Latin square,\\Section~\ref{sec:prompts})};
\node[box] (ft) at (5,0) {2 encoders fine-tuned on\\each training set $\times$ 3 seeds,\\tested on every test set};
\node[box] (pub) at (5,-1.95) {2 public encoders\\without training};
\node[box] (pred) at (10,0) {per-text predictions\\of every system,\\replicate and test set};
\node[box] (an) at (15,0.9) {paired group bootstrap:\\4 questions per comparison\\(H1, H2, H3)};
\node[box] (h4) at (15,-0.9) {variant and position\\tests per system (H4)};
\draw[->] (data) -- (zs); \draw[->] (data) -- (ft); \draw[->] (data) -- (pub);
\draw[->] (zs) -- (pred); \draw[->] (ft) -- (pred); \draw[->] (pub) -- (pred);
\draw[->] (pred) -- (an); \draw[->] (pred) -- (h4);
\end{tikzpicture}
\end{adjustbox}
\caption{Design of the study. All systems are applied to the same test texts; the fine-tuned encoders give in-domain scores (trained and tested on the same dataset) and out-of-domain scores (trained on another dataset). Every number in the main analysis is computed from the stored per-text predictions of the main runs. The four questions and the H4 tests are defined in Section~\ref{sec:stats}; the complementary runs, not shown, are described in Section~\ref{sec:complementary}.}
\label{fig:design}
\end{figure*}

Figure~\ref{fig:design} summarises the design of the study, and Table~\ref{tab:systems} lists all the systems compared in the paper, with the names used throughout. The subscripts of the fine-tuned encoders denote the training condition: \emph{in}, trained on the training split of the test dataset; \emph{out}, trained on another dataset; \emph{zs}, the public checkpoint without any training (Section~\ref{sec:encoders}).

\begin{table*}[!htb]
\centering
\small
\begin{adjustbox}{max width=\linewidth}
\begin{tabular}{llllc}
\toprule
Name in the paper & Family & Trained on our data? & Replicates per test set & Test sets \\
\midrule
Jev & Decision model (API) & No & 4 prompt variants & 10 \\
OpenAI Decisions & Decision model (API) & No & 4 prompt variants & 10 \\
Strands Decider 2B & Decision model (local) & No & 4 prompt variants & 10 \\
GPT-6 Luna & General-purpose LLM (API) & No & 4 prompt variants & 10 \\
Qwen3.5-2B & General-purpose LLM (local) & No & 4 prompt variants & 10 \\
\midrule
tabularisai\textsubscript{in} & Fine-tuned encoder & Yes, on the test dataset & 3 seeds & 10 \\
tabularisai\textsubscript{out} & Fine-tuned encoder & Yes, on another dataset & 27 models (9 source datasets $\times$ 3 seeds) & 10 \\
tabularisai\textsubscript{zs} & Public encoder & No & 1 & 10 \\
XLM-EMO\textsubscript{in} & Fine-tuned encoder & Yes, on the test dataset & 3 seeds & 8 \\
XLM-EMO\textsubscript{out} & Fine-tuned encoder & Yes, on another dataset & 21 models (7 source datasets $\times$ 3 seeds) & 8 \\
XLM-EMO\textsubscript{zs} & Public encoder & No & 1 & 8 \\
\midrule
\multicolumn{5}{l}{\emph{Complementary runs only (Section~\ref{sec:complementary})}} \\
Qwen3.5-2B, label scoring & General-purpose LLM (local) & No & 4 prompt variants & 10 \\
Qwen3.5-4B, generative and label scoring & General-purpose LLM (local) & No & 4 prompt variants & 10 \\
\bottomrule
\end{tabular}
\end{adjustbox}
\caption{Systems compared in the paper. Subscripts: \textsubscript{in}, fine-tuned on the training split of the test dataset; \textsubscript{out}, fine-tuned on a different dataset and averaged over all such source datasets; \textsubscript{zs}, the public checkpoint used without any training on our data. The score of a system with several replicates is the mean over its replicates. XLM-EMO is evaluated on 8 test sets because its starting checkpoint was trained on SemEval-2018. The systems of the last block appear only in the complementary analyses. In tables, Strands Decider 2B is abbreviated as Strands 2B.}
\label{tab:systems}
\end{table*}

\subsection{Task and datasets}
\label{sec:task}
\label{sec:data}

As anticipated in Section~\ref{sec:intro}, the task is to assign each text exactly one of four emotions: \emph{joy}, \emph{sadness}, \emph{anger} or \emph{fear}. These are the four classes used by XLM-EMO, which allows a direct comparison with that model. Texts annotated with more than one of these emotions were removed from all datasets (5{,}237 texts, mostly from SemEval-2018), so that every test text has a single gold label. Quality is measured with macro-F1 over the four classes, computed separately on the test split of each dataset. Macro-F1 gives the same weight to each class, which matters here because the class distribution is strongly unbalanced in several datasets. We report macro-F1 on a 0--1 scale and express differences either directly or in points, where one point equals 0.01 macro-F1. Besides macro-F1, we report the F1 of each class, the latency of every request (Section~\ref{sec:latency}) and, for the systems that return a probability for every class, how well these probabilities are calibrated: the expected calibration error, with 15 equal-width bins on the top probability \citep{guo2017calibration}, and the multiclass Brier score \citep{brier1950verification} (Section~\ref{sec:res-calibration}). The cost of the API-based systems is computed from the token counts that the APIs report and from their list prices (Section~\ref{sec:res-h3}). As a complementary analysis we also recalibrate the probabilities with temperature scaling \citep{guo2017calibration}, which rescales the logits by a single calibration temperature without changing the predicted class, and, for the encoders, remove the effect of the class weights of their loss; both procedures and the formulas of the two calibration measures are given in \statref{}.

We use 10 public datasets in five languages, listed in Table~\ref{tab:data}. In total they contain 66{,}619 texts: 41{,}000 in English, 8{,}864 in German, 8{,}526 in Spanish, 4{,}751 in French and 3{,}478 in Italian. The three splits contain 66{,}352 of them: 267 EmoEvent texts whose original split could not be recovered belong to none (Appendix~\ref{app:data}). The texts come from tweets (EmoEvent, SemEval-2018 Task~1, EMit), Reddit posts (GoEmotions, BRIGHTER), Facebook posts (Universal Joy) and online comments (MultiEmotions-It), so the collection covers different platforms, registers and annotation schemes.

Each dataset was reduced to the four target classes. For GoEmotions we applied the official mapping of its fine-grained labels to Ekman's basic emotions \citep{ekman1992argument}, removed the emotions outside our four classes (disgust, surprise and neutral) and kept the texts with a single remaining emotion. For EMit we kept the texts labelled with exactly one of the four classes, without mapping \emph{love} to \emph{joy}, and we did not use its out-of-domain test set of advertising comments. For Universal Joy we removed duplicates and discarded the class \emph{anticipation}. The resources we excluded and the reasons are listed in Appendix~\ref{app:data}.

We use the official training, validation and test splits wherever they exist (BRIGHTER, GoEmotions, SemEval-2018, EMit and EmoEvent; for EmoEvent the split had to be reconstructed, see Appendix~\ref{app:data}). For MultiEmotions-It and Universal Joy, which have no official split, we created 80/10/10 splits stratified by label; for EMit, which has an official training and test split but no validation split, we held out about 10\% of the training data for validation. All results reported in this paper are computed on the test splits, which together contain 12{,}633 texts.

Every text carries a normalised-text key, obtained by lower-casing the text and removing digits, punctuation, symbols and emoji, so that duplicates and retweets share the same key. The splits we created keep all copies of a text on the same side, and the bootstrap described in Section~\ref{sec:stats} resamples groups of identical texts rather than individual rows, because copies of the same text are not independent observations.

\begin{table*}[t]
\centering
\small
\begin{adjustbox}{max width=\linewidth}
\begin{tabular}{lllrrrr}
\toprule
Dataset & Lang. & Source & Train & Validation & Test & Fear \% \\
\midrule
BRIGHTER \citep{muhammad-etal-2025-brighter} & de & Reddit & 1{,}388 & 202 & 2{,}862 & 7.9 \\
EMit \citep{araque-etal-2023-emit} & it & tweets & 1{,}432 & 159 & 256 & 4.7 \\
EmoEvent \citep{plaza-del-arco-etal-2020-emoevent} & en & tweets & 1{,}995 & 290 & 553 & 5.0 \\
EmoEvent \citep{plaza-del-arco-etal-2020-emoevent} & es & tweets & 2{,}569 & 376 & 725 & 2.5 \\
GoEmotions \citep{demszky-etal-2020-goemotions} & en & Reddit & 24{,}802 & 3{,}118 & 3{,}104 & 2.3 \\
MultiEmotions-It \citep{sprugnoli-2020-multiemotions} & it & comments & 1{,}305 & 163 & 163 & 3.2 \\
SemEval-2018 Task 1 \citep{mohammad-etal-2018-semeval} & en & tweets & 4{,}257 & 572 & 2{,}149 & 9.0 \\
SemEval-2018 Task 1 \citep{mohammad-etal-2018-semeval} & es & tweets & 2{,}368 & 473 & 1{,}908 & 9.6 \\
Universal Joy \citep{lamprinidis-etal-2021-universal} & de & Facebook & 3{,}532 & 441 & 439 & 0.2 \\
Universal Joy \citep{lamprinidis-etal-2021-universal} & fr & Facebook & 3{,}800 & 477 & 474 & 0.5 \\
\midrule
Total & 5 languages & & 47{,}448 & 6{,}271 & 12{,}633 & \\
\bottomrule
\end{tabular}
\end{adjustbox}
\caption{Datasets after reduction to four classes and single-label texts (66{,}619 texts in total). Fear \%: share of \emph{fear}, the least frequent class in every dataset, among all texts of the dataset (all splits). EmoEvent texts whose original split could not be recovered (160 en, 107 es) are in no split, which is why the three splits sum to 66{,}352. The BRIGHTER test split contains each text twice (Appendix~\ref{app:data}).}
\label{tab:data}
\end{table*}

\subsection{Zero-shot systems}
\label{sec:systems}

We evaluate five systems without any training on our data: three decision models and two general-purpose LLMs. They are zero-shot with respect to our task and datasets, not in an absolute sense: their training data are unknown, and Strands Decider 2B obtains its probabilities from a read-out head trained on data that are not documented.

\paragraph{Decision models.} Jev, by TypeSafe AI, is accessed through a closed API (model \texttt{jev-latest}) whose architecture is not disclosed \citep{typesafe-jev}. OpenAI Decisions is the \texttt{/v1/decisions} endpoint of OpenAI, which we ran on the \texttt{gpt-6-luna} model \citep{openai-decisions}. Strands Decider 2B (version \texttt{hobson-v21}) is an open model that, according to its model card, adds to Qwen3.5-2B-Base, the pre-trained model before instruction tuning, a LoRA adapter, a small set of additional trainable weights, and a read-out head, a small layer that turns the internal representation of the model into one probability per option. The adapter was trained in part by distillation from Qwen3.5-4B (Section~\ref{sec:related}); we ran the model on a single cloud GPU \citep{strands-decider}. The three decision models receive a structured question that contains the instructions, the four options and a short description of each option, and they return a probability for every option. The predicted class is the most probable option. Table~\ref{tab:example} shows schematically what the two kinds of zero-shot system receive and return.

\begin{table}[t]
\centering
\small
\begin{adjustbox}{max width=\linewidth}
\begin{tabular}{p{0.47\linewidth}p{0.47\linewidth}}
\toprule
Decision model & General-purpose LLM \\
\midrule
\emph{Request}, structured: instructions with the question (\textit{Quale emozione principale esprime questo messaggio?}); the options \texttt{joy}, \texttt{sadness}, \texttt{anger}, \texttt{fear}, each with a short description; the text to classify. &
\emph{Prompt}, a single string: the same question; the list of options as ``\texttt{- class: description}'' lines in the same order; the instruction to answer with the name of the class only; the text to classify. \\
\addlinespace
\emph{Response}: a probability for each of the four options. The predicted class is the most probable one. &
\emph{Response}: free text, for example ``\texttt{joy}''. Valid only if it names exactly one of the four classes; no probabilities. \\
\bottomrule
\end{tabular}
\end{adjustbox}
\caption{Schematic input and output of the two kinds of zero-shot system for one text (prompt variant 1). The content is identical; only the format differs. Field names of the APIs are omitted.}
\label{tab:example}
\end{table}

\paragraph{Decision models and label scoring.} For Jev and OpenAI Decisions the mechanism is not documented, so we cannot tell whether they differ internally from the label-scoring techniques reviewed in Section~\ref{sec:related}; Strands Decider 2B obtains its probabilities from a trained read-out head. In the complementary runs we also apply label scoring to Qwen3.5-2B and Qwen3.5-4B; for GPT-6 Luna it is not possible, because its API returns the probability of the chosen answer only.

\paragraph{General-purpose LLMs.} GPT-6 Luna was accessed through the OpenAI Responses API \citep{openai-gpt6luna} with reasoning disabled, so that its latency and token use are comparable with the other systems, and with at most 64 output tokens. It was called without temperature or seed: the Responses API accepts a temperature but rejects a seed, and the Decisions endpoint rejects both, so repeated calls can return different answers. We kept the default temperature so that GPT-6 Luna is used under the same conditions as OpenAI Decisions, whose endpoint accepts no temperature, and as the systems are offered. Qwen3.5-2B \citep{qwen35} was run on a single cloud GPU with greedy decoding (always choosing the most probable next token), with its reasoning mode disabled and at most 16 new tokens, which makes it deterministic. The two LLMs receive the same question, options and descriptions as the decision models, serialised into a single prompt that ends by asking for the name of the class only. For all systems, an answer is considered valid only if it names exactly one of the four classes. Answers that contain no class or more than one class are counted as errors rather than discarded. This happened 4 times for OpenAI Decisions (empty answers), 6 times for Qwen3.5-2B and 9 times for GPT-6 Luna, out of 50{,}532 requests per system.

\paragraph{Matched pairs.} Two matched pairs of systems allow a partly controlled test of H3. OpenAI Decisions and GPT-6 Luna run on the same underlying model, so their comparison is the closest available test of the effect of using that model through the OpenAI Decisions endpoint rather than as a text generator. GPT-6 Luna was called with reasoning disabled; the documentation of the OpenAI Decisions endpoint describes no reasoning setting, and we used its defaults, so we cannot verify that the two use the same amount of internal computation. Strands Decider 2B and Qwen3.5-2B have the same size, so their comparison contrasts a decision model with a general LLM at equal scale. The two models also start from the same base model, Qwen3.5-2B-Base, the pre-trained model before instruction tuning: Qwen3.5-2B is its version tuned to follow instructions, while Strands Decider 2B adds a decision adapter trained in part by distillation from a larger model. The second pair therefore contrasts two ways of specialising the same base model, although the decision training also transfers knowledge from a larger model.

\paragraph{Latency.}
\label{sec:latency}
Latency was recorded for every request and is summarised by its median over all requests. For the API-based systems it is the wall-clock duration of the successful HTTP request, from sending it to receiving the complete response, measured on the client; it therefore includes the network and is not directly comparable with the latency of locally hosted models. Requests were concurrent, 4 at a time for the decision APIs and 16 for GPT-6 Luna, except for a fixed sample of 200 texts sent one at a time, which the stored records do not identify and which is therefore included in the medians; a controlled measurement with one request at a time is part of the complementary runs (Section~\ref{sec:complementary}). For Strands Decider 2B and Qwen3.5-2B, which ran on single cloud GPUs, latency covers tokenisation and the forward pass or generation of a single request, after an untimed warm-up. The variants of the two models ran on different GPU types (L4, L40S and A10G; Table~\ref{tab:latency}); with English instructions, in the complementary runs, all variants of both models ran on the L4, which gives a comparison on the same GPU. The computational setup and costs are described in Appendix~\ref{app:compute}.

\subsection{Prompt design}
\label{sec:prompts}

The content of the input is identical for all five zero-shot systems: the same text, the same question and the same descriptions of the four classes. Only the format differs, structured for the decision models and serialised as a prompt for the LLMs. The question and the class descriptions are written in Italian for texts in all five languages, so that the input of every request differs only in the text to be classified. As anticipated in Section~\ref{sec:intro}, this reflects the application for which the study was designed, and we evaluate that configuration; complementary runs repeat the comparison with instructions in English and in the language of each text (Section~\ref{sec:complementary}).

To avoid depending on a single formulation, every system was run with four variants of the question (Table~\ref{tab:prompts} in Appendix~\ref{app:prompts}). The variants differ in wording and in the order in which the four options are presented. The orders form a Latin square: each emotion appears exactly once in each of the four positions across the variants, so that any preference for a position is spread evenly over the classes and can be measured directly. No variant was selected using labelled data. Each variant was run once on all 12{,}633 test texts, and, unless stated otherwise, the score of a zero-shot system is the mean of its four variants.

\subsection{Fine-tuned encoders}
\label{sec:encoders}

As fine-tuned references we use two multilingual encoders based on XLM-R base \citep{conneau-etal-2020-unsupervised}, both already adapted to emotions before our fine-tuning. The first is \texttt{multilingual-emotion-classification} by tabularisai \citep{borisov-2026-tabularisai}, trained on synthetic data generated by LLMs in 23 languages with 11 emotion classes. The second is \texttt{xlm-emo-t}, the public XLM-EMO checkpoint \citep{bianchi-etal-2022-xlm}, built on XLM-T and trained on real emotion datasets in 19 languages. Because the training data of XLM-EMO include SemEval-2018, every comparison that involves XLM-EMO is restricted to the other 8 datasets. According to the description of its training data \citep{bianchi-etal-2022-xlm}, none of these 8 datasets is among its sources: its Italian data come from FEEL-IT, its French and German data are machine translations of Spanish tweets, and EmoEvent and MultiEmotions-It are used there only as out-of-training tests. The only indirect overlap is that its Hindi data are machine translations of 2{,}000 GoEmotions examples.

For each checkpoint we replaced the classification head with a new head for our four classes and fine-tuned the model on the training split of each dataset separately. We followed the recommendations of \citet{mosbach2021stability} and \citet{zhang2021revisiting} for stable fine-tuning: AdamW with a learning rate of $2\cdot10^{-5}$, 10\% linear warm-up followed by linear decay, weight decay 0.01, batch size 32, maximum sequence length 128 and a cross-entropy loss in which each example is weighted by the inverse frequency of its class in the training split, $w_c = N/(4\,n_c)$, where $n_c$ is the number of training texts of class $c$ and $N$ their total. Training lasted up to 20 epochs, with one evaluation per epoch on the validation split and early stopping after five evaluations without improvement in validation macro-F1, restoring the best checkpoint; the selection uses the unweighted validation macro-F1. Each configuration was trained with three seeds (42, 43 and 44), giving 30 models for tabularisai and 24 for XLM-EMO.

Every fine-tuned model was then tested on the test split of every dataset, which yields a ``trained on $\rightarrow$ tested on'' matrix. The diagonal of this matrix gives the \emph{in-domain} performance, marked with the subscript \emph{in} (for example, tabularisai\textsubscript{in}): the model is trained on the same dataset on which it is tested, and its score is the mean of the three seeds. The off-diagonal cells give the \emph{out-of-domain} performance, marked with the subscript \emph{out}: for each test set, we average all the models trained on a different dataset, that is, 9 source datasets for tabularisai and 7 for XLM-EMO, each with three seeds. Note that \emph{out} includes models trained on the same corpus in another language, such as Universal Joy French for Universal Joy German, which are out-of-dataset but not fully out-of-domain. As an additional, exploratory reference we also evaluated both public checkpoints without any training on our data, marked with the subscript \emph{zs}, restricting their output to the four target classes by a softmax over the logits of these classes only (the other seven classes of tabularisai are ignored).

\subsection{Statistical analysis}
\label{sec:stats}

The analysis plan fixed the equivalence margin, the families of comparisons (except the later inclusion of Qwen3.5-2B in the H1 family) and the classification of the primary outcome before any paired test was computed (Section~\ref{sec:intro}); the verdict rules for H3 and H4 and the complementary analyses were defined afterwards (Appendix~\ref{app:deviations}). All tests use the stored per-text predictions.

\paragraph{Paired group bootstrap.} Uncertainty is estimated with a bootstrap \citep{efron1993introduction} that resamples groups of identical texts within each test set ($B=10{,}000$). The same resampled texts are used for all systems, so that every difference is paired \citep{berg-kirkpatrick-etal-2012-empirical}; details are in \statref{}.

\paragraph{Multiplicity.} Comparisons are grouped into one family per hypothesis, and Holm's procedure \citep{holm1979simple} is applied within each family, separately for each question (for Question~2, over all pairs and test sets of the family); adjusted $p$-values are capped at 1. Table~\ref{tab:families} lists, for each hypothesis, the comparisons and the rule that turns their outcomes into a verdict. The H3 family consists of the two matched pairs of Section~\ref{sec:systems} and of the unmatched pair Jev $-$ GPT-6 Luna; H4 is tested separately for each system.

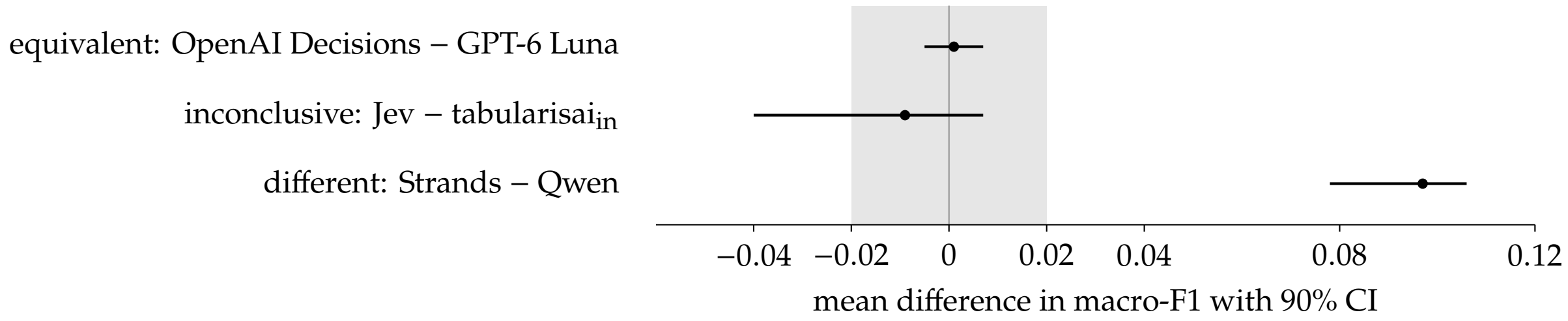
\begin{figure}[t]
\centering
\begin{adjustbox}{max width=\linewidth}
\begin{tikzpicture}[x=50cm,y=0.7cm,font=\small]
\fill[black!10] (-0.02,0.4) rectangle (0.02,3.6);
\draw[black!40] (0,0.4) -- (0,3.6);
\draw (-0.06,0.4) -- (0.12,0.4);
\foreach \v/\t in {-0.04/$-0.04$,-0.02/$-0.02$,0/$0$,0.02/$0.02$,0.04/$0.04$,0.08/$0.08$,0.12/$0.12$}
  \draw (\v,0.4) -- (\v,0.3) node[below] {\t};
\draw[thick] (-0.005,3) -- (0.007,3); \fill (0.001,3) circle (1.5pt);
\node[anchor=east] at (-0.065,3) {equivalent: OpenAI Decisions $-$ GPT-6 Luna};
\draw[thick] (-0.040,2) -- (0.007,2); \fill (-0.009,2) circle (1.5pt);
\node[anchor=east] at (-0.065,2) {inconclusive: Jev $-$ tabularisai\textsubscript{in}};
\draw[thick] (0.078,1) -- (0.106,1); \fill (0.097,1) circle (1.5pt);
\node[anchor=east] at (-0.065,1) {different: Strands $-$ Qwen};
\node[below] at (0.03,-0.35) {mean difference in macro-F1 with 90\% CI};
\end{tikzpicture}
\end{adjustbox}
\caption{The three outcomes of the primary question, illustrated with three comparisons from this study. The grey band is the equivalence margin $\pm0.02$. A 90\% CI inside the band means \emph{equivalent}; a Holm-adjusted $p<0.05$ (a 95\% CI that excludes zero, before correction) means \emph{different}; otherwise the outcome is \emph{inconclusive}, as when the CI crosses both zero and the edge of the band.}
\label{fig:outcomes}
\end{figure}

\paragraph{Four questions per comparison.} For each pair of systems $A$ and $B$, with $\Delta = \text{F1}_A - \text{F1}_B$, Table~\ref{tab:questions} lists four questions: the mean over test sets (primary), each test set, the consistency of the sign across test sets, and the difference to be expected on a new dataset. The first three were specified in the plan; the fourth, answered with a random-effects meta-analysis, was added after the first results. The primary outcome is classified as equivalent, different or inconclusive by combining the two one-sided tests procedure (TOST) \citep{schuirmann1987comparison,lakens2017equivalence}, with margin $\delta=0.02$, with the usual test of difference (Table~\ref{tab:questions}, Figure~\ref{fig:outcomes}); we also report the outcomes for $\delta=0.01$ and $\delta=0.03$.

\begin{table*}[t]
\centering
\small
\begin{adjustbox}{max width=\linewidth}
\begin{tabular}{p{3.2cm}p{4.2cm}p{4.4cm}p{4.6cm}}
\toprule
Question & Quantity & Method & How to read it \\
\midrule
1. How do $A$ and $B$ compare on average? (primary) & Unweighted mean of $\Delta$ over test sets & Paired bootstrap CI and $p$-value; TOST with margin $\delta$ & Checked in this order: \emph{equivalent} if the 90\% CI lies within $[-\delta, +\delta]$; \emph{different} if $p<0.05$ after Holm's correction for multiple comparisons; otherwise \emph{inconclusive}, i.e.\ compatible both with a negligible difference and with one larger than the margin \\
2. Does $A$ differ from $B$ on a given test set? & $\Delta$ on each test set & Paired bootstrap: 95\% CI and $p$-value, Holm correction & Significantly better, not different, or significantly worse on that test set \\
3. Is the sign of $\Delta$ consistent across test sets? & Observed $\Delta$ of each test set & Exact Wilcoxon signed-rank test \citep{wilcoxon1945individual,demsar2006statistical}; reported in Table~\ref{tab:wilcoxon} & A significant result means that one system tends to win across datasets \\
4. What can be expected on a new dataset? & Mean $\Delta$ in a population of datasets & Random-effects meta-analysis: CI and 95\% prediction interval & The prediction interval gives the plausible range of $\Delta$ on a dataset not in this study \\
\bottomrule
\end{tabular}
\end{adjustbox}
\caption{The four questions asked for each comparison between two systems $A$ and $B$, and how the outcome of each is read.}
\label{tab:questions}
\end{table*}

\paragraph{Margin.} The margin $\delta=0.02$ is close to the mean spread between the best and the worst of three seeds of the fine-tuned encoders (0.018 for tabularisai and 0.021 for XLM-EMO), a correspondence checked after the margin was fixed. The spread is measured on single datasets, whereas the margin applies to the mean over 8 or 10 datasets, so the correspondence holds as an order of magnitude.

\paragraph{Precision of H1.} Equivalence can be shown only when the 90\% CI of $\Delta$ is narrower than the band (0.04). This holds for systems whose predictions are strongly correlated, such as OpenAI Decisions and GPT-6 Luna (CI width 0.012), but not for a zero-shot system compared with a fine-tuned encoder, which agree on far fewer texts: their 90\% CIs are 0.045--0.058 wide, so for H1 equivalence within $\pm0.02$ was out of reach whatever the observed difference. Even with an observed difference of exactly zero, equivalence could have been shown only with margins of at least half that width, about 0.023 to 0.029. With 80\% power it would have required test sets about 4--7 times larger if the true difference were zero, and 9--23 times larger at the observed differences (\statref{}). We therefore report for each pair the smallest equivalence margin $\delta_{\min}$, the larger absolute bound of the 90\% CI: the TOST concludes equivalence for any margin larger than $\delta_{\min}$. It is a descriptive transformation of the CI, not a new margin chosen after the fact, and readers can compare it with the margin they consider negligible.

\begin{table*}[t]
\centering
\small
\begin{adjustbox}{max width=\linewidth}
\begin{tabular}{>{\raggedright\arraybackslash}p{2.6cm}>{\raggedright\arraybackslash}p{6.2cm}>{\raggedright\arraybackslash}p{6.8cm}}
\toprule
Hypothesis & Comparisons (one Holm family per hypothesis) & Verdict \\
\midrule
H1, in-domain & each zero-shot system vs.\ tabularisai\textsubscript{in} and XLM-EMO\textsubscript{in} (10 pairs) & \emph{supported} if the primary outcome is \emph{equivalent}; \emph{inconclusive} if it is neither equivalent nor different; systems that are significantly worse are reported as \emph{worse} \\
H2a, loss out of domain & in-domain minus out-of-domain score of each encoder & \emph{supported} if the loss is significant \\
H2b, zero-shot vs.\ out of domain & three leading systems and Strands Decider 2B vs.\ tabularisai\textsubscript{out} and XLM-EMO\textsubscript{out} (10 pairs in one family, including the two H2a comparisons) & \emph{supported} if the zero-shot system is significantly better; \emph{generalisable to new datasets} if also under the random-effects model \\
H3, decision vs.\ LLM & OpenAI Decisions vs.\ GPT-6 Luna and Strands Decider 2B vs.\ Qwen3.5-2B (matched); Jev vs.\ GPT-6 Luna (3 pairs) & \emph{supported} if the decision model is equivalent or significantly better in quality and has a lower median latency (only the two matched pairs determine the verdict); cost described, not tested \\
H4, prompt & each zero-shot system separately (no pairs) & \emph{supported} if the upper bound of the 90\% CI of the mean SD is below 0.02; \emph{inconclusive} if the mean SD is below 0.02 but the upper bound of its 90\% CI is not; \emph{rejected} if the mean SD exceeds it. Permutation and position tests are descriptive \\
\midrule
Exploratory & the two encoders against each other; zero-shot systems vs.\ the public checkpoints (\textsubscript{zs}) & no verdict (\ifdefined\mdpiversion Table~\ref{tab:exploratory}\else Appendix~\ref{app:extra}\fi) \\
\bottomrule
\end{tabular}
\end{adjustbox}
\caption{Hypotheses, comparisons and verdict rules. Outcomes refer to the primary question (Table~\ref{tab:questions}) unless stated otherwise.}
\label{tab:families}
\end{table*}

\paragraph{On our datasets and on new datasets.} A result holds \emph{on our datasets} when the mean over our test sets is significant. To estimate what can be expected on a new dataset, a random-effects meta-analysis treats the test sets as a sample from a population of datasets \citep{higgins2009reevaluation}; with the between-dataset variance $\tau^2$ estimated by restricted maximum likelihood (REML) and the CI computed with the Hartung--Knapp method (\statref{}), it gives a CI for the population mean, a 95\% prediction interval for the difference on a new dataset, and the heterogeneity $I^2$, the share of the variation between datasets that is not due to chance \citep{higgins2002quantifying}. A result that is also significant under this model is called \emph{generalisable to new datasets}. Unlike the CI of the mean, the prediction interval does not shrink as datasets are added, because it includes the variability between datasets (\statref{}).

\paragraph{Sensitivity to the prompt (H4).} For each system, the standard deviation (SD) of macro-F1 across the four prompt variants is averaged over the test sets; the system is \emph{stable} if the upper bound of the bootstrap 90\% CI of this mean is below $\delta$. This criterion alone determines the verdict on H4. We use the same margin as for the other hypotheses because it expresses, on the same macro-F1 scale, the smallest change that matters in practice; here it bounds the typical deviation of a system from its own mean across formulations rather than a difference between two systems. A permutation test of the variant effect and the position bias, measured against the 25\% share expected under the Latin square, are reported descriptively (\statref{}).

\subsection{Complementary runs}
\label{sec:complementary}

After the first results, five sets of runs, grouped in the four rows of Table~\ref{tab:complementary-design}, were added between 28 October and 2 November 2026 to test points that the main design left open (Appendix~\ref{app:deviations}). They use the same texts and bootstrap resamples as the main analysis, so their comparisons are paired with it; they are reported as complementary analyses and do not change any verdict. Table~\ref{tab:complementary-design} summarises them. In label scoring, each of the four labels is scored by its log-probability as the continuation of the generative prompt, formatted with the chat template of the model with its reasoning mode disabled and tokenised together with the label, summed over the tokens of the label or averaged per token; the four scores are normalised with a softmax to give probabilities. The translations of the instructions were produced with an LLM, Claude Sonnet 5.5, checked by one of the authors, and not reviewed by native speakers. These runs requested Jev as \texttt{jev-1.13.0}, the version served by the alias \texttt{jev-latest} in the main runs, and used the same model commits for the open models. With two runs of each variant, the variation of macro-F1 across variants can be split into two parts: the run-to-run variation, estimated from the differences between the two runs of the same variant, and the variation due to the prompt, estimated from the differences between variants after removing the share expected from run-to-run variation. Both are computed on each test set, summarised as SDs averaged over test sets and given with bootstrap CIs (\statref{}).

\begin{table*}[t]
\centering
\small
\begin{adjustbox}{max width=\linewidth}
\begin{tabular}{>{\raggedright\arraybackslash}p{2.4cm}>{\raggedright\arraybackslash}p{3.6cm}>{\raggedright\arraybackslash}p{4.6cm}>{\raggedright\arraybackslash}p{3.6cm}>{\raggedright\arraybackslash}p{2.2cm}}
\toprule
Run & Systems & What changes & Question & Results \\
\midrule
Label scoring & Qwen3.5-2B, Qwen3.5-4B & Probabilities of the four labels read from the generative prompt (summed and averaged log-probabilities); Qwen3.5-4B, from which Strands Decider 2B was partly distilled, also run generatively & Does the advantage of Strands Decider 2B come from the decision format or from its training? & Section~\ref{sec:res-h3}, Table~\ref{tab:complementary} \\
Language of the instructions & Strands Decider 2B and Qwen3.5-2B (English and text language); Jev, OpenAI Decisions, GPT-6 Luna and Qwen3.5-4B (text language) & Question and class descriptions translated into English and, for the German, Spanish and French test sets, into the language of the text & Do the results depend on the Italian instructions? & Section~\ref{sec:res-h4}, Tables~\ref{tab:complementary} and~\ref{tab:complementary-h4} \\
Controlled latency & Jev, OpenAI Decisions, GPT-6 Luna & 600 texts (60 per test set, variant 1), one request at a time, the three APIs in random order, two time windows; ratio of medians with a bootstrap CI over texts & Is the latency ratio an artefact of concurrency? & Section~\ref{sec:res-h3}, Table~\ref{tab:controlled-latency} \\
Second run & Jev, GPT-6 Luna & The four variants run a second time & Is the variant effect due to the prompt or to run-to-run variation? & Section~\ref{sec:res-h4}, Tables~\ref{tab:complementary} and~\ref{tab:complementary-h4} \\
\bottomrule
\end{tabular}
\end{adjustbox}
\caption{The complementary runs: systems, what changes with respect to the main runs, the question each run answers, and where its results are reported.}
\label{tab:complementary-design}
\end{table*}

\section{Results}
\label{sec:results}

The results are reported in the order of the hypotheses. Section~\ref{sec:overview} gives an overview and introduces Table~\ref{tab:verdicts}, which summarises the outcome of every hypothesis for every system. Section~\ref{sec:res-h1} follows the four questions of Table~\ref{tab:questions}; Sections~\ref{sec:res-h2}--\ref{sec:res-h4} report the primary outcome and add the other questions where they matter, with the complementary runs where they bear on each hypothesis; and Section~\ref{sec:res-calibration} describes the calibration of the probabilities.

\subsection{Overview}
\label{sec:overview}

Table~\ref{tab:overview} reports the macro-F1 of every system on every test set. Averaged over the 10 test sets, the three leading zero-shot systems obtain almost the same score: 0.708 for Jev (95\% CI [0.686, 0.724]), 0.707 for OpenAI Decisions and 0.706 for GPT-6 Luna. Strands Decider 2B reaches 0.669 and Qwen3.5-2B 0.571. The encoder fine-tuned in-domain, tabularisai\textsubscript{in}, obtains 0.717 (95\% CI [0.702, 0.744]), and 0.553 out-of-domain; on its 8 test sets, XLM-EMO obtains 0.690 in-domain and 0.596 out-of-domain. The public checkpoints used without any training score 0.525 (tabularisai) and 0.616 (XLM-EMO). The difficulty of the datasets varies widely: the three leading zero-shot systems score between 0.78 and 0.85 on EmoEvent and SemEval-2018, whereas no system exceeds 0.59 on Universal Joy, where the class \emph{fear} is almost absent. Table~\ref{tab:verdicts} is the map of the rest of this section: for each system and hypothesis it gives the outcome in bold and, below it, the numbers on which it rests, in macro-F1 points.

\input{tables/overview}

\input{tables/verdicts_numeric}

\subsection{H1: zero-shot systems versus in-domain fine-tuning}
\label{sec:res-h1}

\begin{figure*}[t]
\centering
\includegraphics[width=\textwidth]{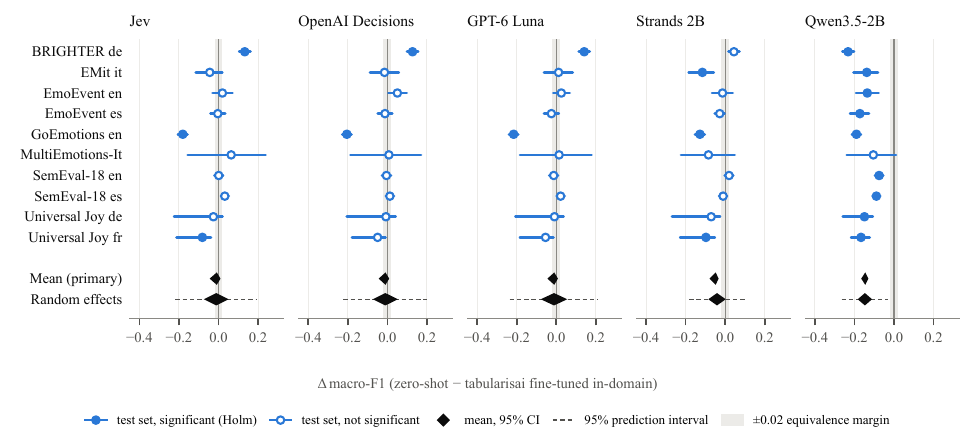}
\caption{H1: macro-F1 difference between each zero-shot system and tabularisai fine-tuned in-domain, per test set (group-bootstrap 95\% CI; filled markers are significant after Holm correction). The bottom rows give the mean over test sets (primary analysis) and the random-effects mean, with its 95\% prediction interval for a new dataset as a dashed line. The grey band is the $\pm0.02$ equivalence margin.}
\label{fig:h1}
\end{figure*}

H1 is inconclusive for the three leading zero-shot systems, an outcome determined largely by the precision of the comparison (Section~\ref{sec:stats}). On average they are about one point below in-domain fine-tuning, and their smallest equivalence margin $\delta_{\min}$ is 0.039 to 0.049 (Table~\ref{tab:h1}). Strands Decider 2B and Qwen3.5-2B are significantly worse. Figure~\ref{fig:h1} shows the per-dataset differences with respect to tabularisai.

\input{tables/h1}

\paragraph{Question 1: mean over test sets (primary).} For the primary comparison, Jev minus tabularisai\textsubscript{in}, the mean difference is $-0.009$, not significant, with a 90\% CI of $[-0.040, 0.007]$: the data exclude an advantage of the zero-shot system larger than two points ($p=0.002$) but not a deficit of that size ($p=0.37$), so the outcome is \emph{inconclusive}; it stays inconclusive with a margin of three points. OpenAI Decisions and GPT-6 Luna, and the comparison with XLM-EMO\textsubscript{in} on 8 test sets, give practically the same values, so the conclusion depends neither on the zero-shot system nor on the reference encoder.

\paragraph{Questions 2 and 3: individual test sets and consistency.} The small average difference hides large differences in both directions (Figure~\ref{fig:h1}; Table~\ref{tab:h1-per-dataset}\ifdefined\mdpiversion{} in the Supplementary Materials\fi). The three leading systems are significantly better than either fine-tuned encoder on BRIGHTER, by 13 to 16 points, and significantly worse on GoEmotions, by 17 to 22 points; Jev is also significantly worse than tabularisai\textsubscript{in} on Universal Joy French. On the other datasets the differences are not significant after correction, and equivalence within two points on a single dataset is established only for Jev on SemEval-2018 English (uncorrected TOST $p=0.047$). Accordingly, the Wilcoxon test finds no consistent sign of the difference for the three leading systems, whereas Qwen3.5-2B is consistently worse against tabularisai\textsubscript{in} (Table~\ref{tab:wilcoxon}).

\paragraph{Question 4: a new dataset.} The random-effects estimate for Jev, $-0.011$, is close to the simple mean, but the heterogeneity between datasets is very large ($I^2=0.97$, $\tau=0.085$: the true difference typically varies by about $\pm9$ points from one dataset to another), and the 95\% prediction interval for a new dataset is $[-0.22, 0.20]$: on a dataset not included in this study, either approach could win by a wide margin. Under this model the deficit of Strands Decider 2B is no longer significant.

\paragraph{Robustness of H1 (complementary).} The verdict on H1 does not depend on how the test sets are aggregated, but the size of the deficit does: weighting them by the number of test texts, which favours GoEmotions and BRIGHTER, puts the fine-tuned encoder 1.3 to 2.2 points ahead with CIs that exclude zero, but every aggregation except the fixed-effect estimate, inappropriate given the heterogeneity, places the three leading systems between 0 and 2.6 points below in-domain fine-tuning (Table~\ref{tab:aggregation}). The statement ``no detectable difference'' therefore depends on how the datasets are weighted, whereas ``on average between zero and about two points below in-domain fine-tuning'' does not. Without the two Universal Joy test sets, where \emph{fear} has only 1 and 2 texts and its F1 is all-or-nothing, the three leading systems move to between $-0.5$ and $+0.4$ points and $\delta_{\min}$ falls to 0.024--0.032. No outcome of H1--H3 at $\delta=0.02$ changes; only the equivalence of Jev and GPT-6 Luna within one point no longer holds (Table~\ref{tab:no-uj}). The heterogeneity remains ($I^2=0.98$ for Jev), so it comes from datasets such as BRIGHTER and GoEmotions. The advantage of fine-tuning grows with the size of the training split (Spearman $\rho$ from $-0.62$ to $-0.88$ for the three leading systems), but the relation weakens without GoEmotions ($-0.48$ to $-0.82$) and, with 8--10 datasets and training size confounded with other properties of the datasets, should be read as a hypothesis for future work (Table~\ref{tab:moderator}).

\providecommand{\barrierbeforeHtwo}{\FloatBarrier}
\barrierbeforeHtwo
\subsection{H2: performance outside the training dataset}
\label{sec:res-h2}

\input{tables/h4}

H2 is supported. H2a holds for both encoders, which lose much of their quality outside their training dataset; H2b holds clearly against tabularisai and, on our datasets only, against XLM-EMO.

\paragraph{H2a: loss of the fine-tuned encoders.} Out of domain, tabularisai loses 16.3 points and XLM-EMO 9.5 (Table~\ref{tab:h4}), significantly with every test, including the random-effects model. Fine-tuning on another dataset still improves on the public tabularisai checkpoint (0.553 against 0.525; paired difference $+0.028$, 95\% CI $[0.017, 0.040]$). Out-of-domain XLM-EMO also scores on average below the public \texttt{xlm-emo-t} checkpoint used without any training (0.596 against 0.616; paired difference $-0.020$, 95\% CI $[-0.031, -0.009]$, a complementary test without multiplicity correction): after fine-tuning on a different emotion dataset, the model did worse on the target dataset than without fine-tuning (Section~\ref{sec:disc-when}).

\paragraph{H2b: zero-shot systems against out-of-domain encoders.} Against tabularisai\textsubscript{out} the three leading systems and Strands Decider 2B are better on average by 11.5 to 15.5 points, Jev on 8 of the 10 test sets, and the advantage is consistent across test sets and extends to new datasets (Wilcoxon and random-effects tests significant). Against XLM-EMO\textsubscript{out} the advantage is smaller, consistent with its starting checkpoint having been trained on real emotion data in 19 languages, whereas tabularisai was trained on synthetic data. It is 4 to 9 points and significant on our 8 test sets, but on only 1 to 3 of them individually (OpenAI Decisions and GPT-6 Luna are worse on one), and neither the Wilcoxon test nor the random-effects model is significant; the prediction intervals include negative values (for Jev $[-0.13, 0.29]$), so the advantage cannot be generalised to new datasets.

\paragraph{Against the public checkpoints (exploratory).} A practitioner without labelled data could also use a public emotion model as it is. This comparison is less favourable to the zero-shot systems (Table~\ref{tab:exploratory}): the three leading systems are 18 points better than the tabularisai checkpoint, trained on synthetic data, but only about 6 points better than the public XLM-EMO checkpoint, on our datasets only, and Strands Decider 2B is not distinguishable from it.

\paragraph{Sources of the transfer (complementary).} For tabularisai, a source dataset in the same language as the test set gives on average 4.2 points more than a source in another language, a difference that does not hold under the random-effects model; for XLM-EMO the language of the source makes no difference. The best source for a test set is often the same corpus in another language (for example Universal Joy French for Universal Joy German, or EmoEvent Spanish for EmoEvent English), which suggests that the similarity of corpus and annotation matters more than the language. Even against the best source chosen on the test set itself, an oracle choice that favours the fine-tuned encoder, Jev remains ahead on average, by 7.1 points against tabularisai and 4.8 against XLM-EMO, though not under the random-effects model (Table~\ref{tab:h4-extra}).

\FloatBarrier
\subsection{H3: decision models versus general-purpose LLMs}
\label{sec:res-h3}

\input{tables/h2}

H3 is supported for quality and latency. OpenAI Decisions gives the same quality as GPT-6 Luna, the LLM on which it runs, with a median latency about five times lower. Strands Decider 2B clearly outperforms Qwen3.5-2B, a general LLM of the same size; a control with label scoring shows that this advantage comes from its training, not from the decision format. Monetary cost does not separate the systems.

\paragraph{Quality.} OpenAI Decisions and GPT-6 Luna are \emph{equivalent} even within one point ($+0.001$, 90\% CI $[-0.005, 0.007]$), with no significant difference on any test set. Jev is also equivalent to GPT-6 Luna within one point, though narrowly (TOST $p=0.048$); this pair is not matched, since the architecture of Jev is not disclosed, so it shows that the two systems reach the same quality, not why, and gives no verdict on H3. Strands Decider 2B is 9.7 points better than Qwen3.5-2B, on 6 test sets and worse on none, consistently across test sets and also under the random-effects model; for the two equivalent pairs neither the Wilcoxon test nor the random-effects model finds a direction (Table~\ref{tab:h2}).

\paragraph{Latency and cost.} In the main runs the two decision APIs answer in a median of 220--229\,ms per request and GPT-6 Luna in 1{,}107\,ms, with a wider gap in the tail (95th percentiles 273--280 against 3{,}223\,ms; Table~\ref{tab:latency}). Both figures include the network and were measured from the same infrastructure. GPT-6 Luna, however, was called with higher concurrency, which could inflate its latency if the service throttles concurrent calls. We therefore repeated the measurement with one request at a time, calling the three APIs one after the other for each of 600 texts in two time windows: GPT-6 Luna is 5.4 times slower than OpenAI Decisions (95\% CI $[5.1, 5.6]$), with the same ratio in both windows (Table~\ref{tab:controlled-latency}). Absolute latencies depend on the day and the network: both OpenAI endpoints were 70--90\% slower than in the main runs and Jev faster, so Jev, as fast as OpenAI Decisions in the main runs, is 2.5 times faster in the controlled measurement. The token counts reported by the APIs give about USD 0.02 per thousand requests for each API, two of them from prices we could not verify, so cost does not separate the systems. Strands Decider 2B is also faster than Qwen3.5-2B (median 65 against 146\,ms; 62 against 143\,ms with English instructions, with both models on the same L4 GPU), partly by construction, since it reads its probabilities from a single forward pass whereas Qwen3.5-2B generates its answer token by token.

\paragraph{Format or training? (complementary)} Label scoring gives Qwen3.5-2B the output format of a decision model, a probability for every option, without any training. Its quality is equivalent to that of generative Qwen3.5-2B even within one point ($\Delta=+0.001$; Table~\ref{tab:complementary}), Strands Decider 2B remains 9.6 points better, and label scoring leaves the instability and the position bias of Qwen3.5-2B unchanged (Table~\ref{tab:complementary-h4}). It is, however, faster than generation (91 against 146\,ms). At two billion parameters, the advantage of the decision model in quality therefore comes from its training, the decision adapter with its partial distillation from Qwen3.5-4B, and not from the decision format. Qwen3.5-4B shows how much of the teacher this training transfers. Strands Decider 2B is equivalent to it within two points ($\Delta=+0.003$), whether Qwen3.5-4B is used generatively or with label scoring, which are themselves equivalent. It is also about half as sensitive to the prompt (SD of 0.012 against 0.024--0.025) and about three times faster (62 against 178\,ms on the same L4 GPU, Strands Decider 2B with English instructions and Qwen3.5-4B with label scoring; Table~\ref{tab:latency}). Jev remains 4.3 points better than Qwen3.5-4B. The gap between Strands Decider 2B and Qwen3.5-2B persists with instructions in English and in the language of the text (Section~\ref{sec:res-h4}).

\FloatBarrier
\subsection{H4: sensitivity to the prompt}
\label{sec:res-h4}

\input{tables/h3}

\begin{figure*}[t]
\centering
\includegraphics[width=\textwidth]{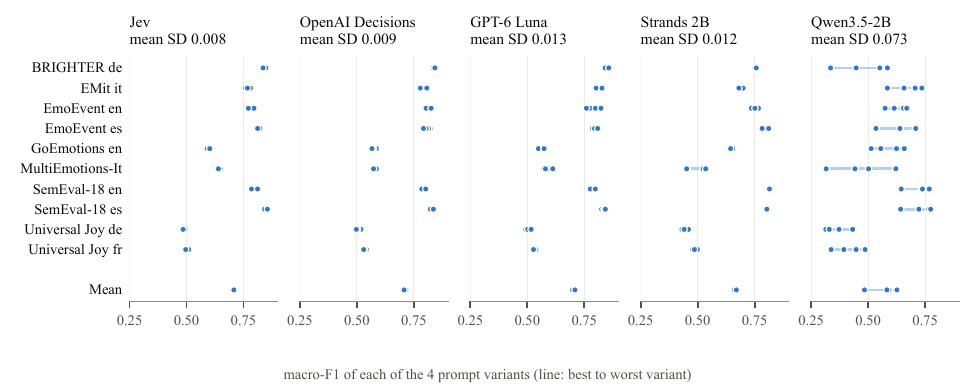}
\caption{H4: macro-F1 of each of the four prompt variants per test set; the line spans the best and worst variant. Mean SD across variants is given under each system name.}
\label{fig:h3}
\end{figure*}

H4 is supported for Jev, OpenAI Decisions and GPT-6 Luna, inconclusive for Strands Decider 2B, which misses the stability threshold by a narrow margin, and rejected for Qwen3.5-2B (Table~\ref{tab:h3}, Figure~\ref{fig:h3}; position bias in Figure~\ref{fig:position}\ifdefined\mdpiversion{} of the Supplementary Materials\fi).

\paragraph{Stability and variant effect.} The mean SD of Jev, OpenAI Decisions and GPT-6 Luna across the four variants is between 0.008 and 0.013, with upper bounds of 0.013 to 0.018, so all three are \emph{stable}. Strands Decider 2B narrowly misses the threshold (upper bound 0.0204), mainly because of MultiEmotions-It, a test set of only 163 texts, where its SD is 0.045; with English instructions it would meet it (upper bound 0.014; Table~\ref{tab:complementary-h4}). Qwen3.5-2B is clearly not stable (mean SD 0.073; on MultiEmotions-It an SD of 0.13 and 31 points between its best and worst variant). The permutation test detects a strong variant effect for Qwen3.5-2B, significant on 9 of the 10 test sets, and a small one for GPT-6 Luna and Jev: significant on one test set each, but with a mean SD well below the margin; none for OpenAI Decisions and Strands Decider 2B.

\paragraph{Prompt or run-to-run variation? (complementary)} Since Jev and GPT-6 Luna are not deterministic and each variant was run at a different time, their small variant effect could come from the run rather than from the prompt. A second run of the four variants gives the same mean quality, equivalent within one point, and the same answer for 98.9\% of the answers (texts $\times$ variants) for Jev and 95.4\% for GPT-6 Luna. The SD due to the prompt is 0.007 and 0.008, and the SD between two runs of the same variant 0.004 and 0.009 (Table~\ref{tab:complementary-h4}): both systems have a real but small prompt effect, for GPT-6 Luna of the same size as its run-to-run variation, and their stability verdicts do not change.

\paragraph{Language of the instructions (complementary).} With English instructions, both 2B models lose less than one point and the gap between them does not change. With instructions in the language of the text, on the German, Spanish and French test sets, Strands Decider 2B is unchanged, whereas Qwen3.5-2B gains 6.2 points, so that its gap from Strands Decider 2B on these test sets falls from 13.0 to 6.4 points. What matters for the small general LLM is therefore that the instructions match the language of the text, not that they are in English. The three leading systems change by less than one point (Jev $+0.006$, OpenAI Decisions $+0.006$, GPT-6 Luna $0.000$), all equivalent to Italian within two points (English instructions were tested only on the 2B models), and the H3 comparison between OpenAI Decisions and GPT-6 Luna is unchanged; OpenAI Decisions refused to answer two French texts with French instructions, counted as errors. Qwen3.5-4B gains 1.5 points, a quarter of the gain of Qwen3.5-2B, and stays equivalent to Strands Decider 2B (Table~\ref{tab:complementary}). With instructions in the language of the text, on these five test sets, the SD of Qwen3.5-2B halves (from 0.075 to 0.036) but it remains far from stable, whereas the three leading systems remain stable (Table~\ref{tab:complementary-h4}).

\paragraph{Position bias.} With about 50{,}000 answers per system the position-bias test is significant for every system, since even half a percentage point is detectable, so the relevant information is the size of the deviation. For the three decision models and GPT-6 Luna, the share of answers given to any position deviates from the expected 25\% by at most 1.4 percentage points of answers. Qwen3.5-2B chooses the first option 10.6 percentage points more often than expected, in about one case in three. With English instructions it also leaves the required format more often (178 invalid answers, against 6 with Italian instructions, counted as errors), and its preference for the first option grows to 17.6 percentage points (17.0 with label scoring), whereas instructions in the language of the text reduce it (from 11.5 to 3.9 percentage points on the German, Spanish and French test sets); Strands Decider 2B stays within 2 percentage points in every condition (Table~\ref{tab:complementary-h4}).

\FloatBarrier
\subsection{Calibration of the probabilities}
\label{sec:res-calibration}

This complementary analysis qualifies one of the advantages of the decision format, the probability it returns for every option. The decision models, the encoders and the Qwen models used with label scoring return a probability for every class; GPT-6 Luna and Qwen3.5-2B used generatively return none. Table~\ref{tab:calibration} gives the calibration of each system and Table~\ref{tab:calibration-tests} the paired bootstrap of the differences.

\paragraph{As returned.} The expected calibration error \citep{guo2017calibration} of the three decision models, 0.088--0.098, is lower than that of the encoders fine-tuned in-domain (0.129 and 0.134) and much lower than out of domain (0.195--0.226), as previously observed for pre-trained transformers outside their training distribution \citep{desai-durrett-2020-calibration}; all these differences have 95\% intervals that exclude zero. The multiclass Brier score \citep{brier1950verification}, which combines calibration and accuracy, differs significantly only out of domain: in-domain, the higher accuracy of the encoders offsets their worse calibration.

\paragraph{After recalibration.} The comparison with the encoders is not neutral, because they are much more confident than they should be. Removing the effect of the class weights of their loss leaves their calibration unchanged, whereas temperature scaling \citep{guo2017calibration} lowers their expected calibration error to 0.047--0.052 in-domain and 0.059 out of domain, and that of the decision models to 0.037--0.047. In-domain the gap then disappears: no difference in expected calibration error is distinguishable from zero, and Strands Decider 2B has a significantly higher Brier score than both encoders. Out of domain the decision models remain better calibrated in five of the six comparisons. Recalibration needs labelled texts from the target domain, which we take from the test sets themselves (\statref{}); without them, a calibration temperature estimated on the in-domain test set of the same model lowers the out-of-domain error only to 0.10--0.12, still above that of the decision models as returned.

\paragraph{Matched comparison.} Strands Decider 2B is better calibrated than label-scored Qwen3.5-2B, built on the same base model, both as returned (0.088 against 0.104; Brier 0.301 against 0.403) and after recalibration (0.037 against 0.058), with significant differences in both measures. Against label-scored Qwen3.5-4B, twice its size, its error as returned is higher (difference 0.013, 95\% CI 0.005--0.022) and the Brier scores do not differ significantly. For large language models, the probabilities of the options can be well calibrated when the format is suitable \citep{kadavath2022language}.

\input{tables/calibration}


\section{Discussion}
\label{sec:discussion}

We discuss first the two main questions of the paper, whether zero-shot systems can replace fine-tuning and what the decision format adds, then the heterogeneity between datasets behind both, and finally robustness to the prompt and what our analysis implies for evaluation practice.

\subsection{When to use a zero-shot system, and when a decision model}
\label{sec:disc-when}

On the first question, the results give a practical answer for a practitioner without labelled data from the target domain. The three leading zero-shot systems, decision models and general LLM alike, are the best option among those we tested. They are clearly better than a fine-tuned encoder borrowed from another dataset, and better by about six points, on our datasets only, than the strongest public emotion model used without training, which is the harder alternative for such a practitioner; on average they come close to in-domain fine-tuning without being demonstrably equivalent to it. When labelled data from the target domain are abundant, fine-tuning may be preferable: it wins clearly on GoEmotions, the dataset with the largest training split (24{,}802 texts), and the exploratory analysis of H1 suggests that its advantage grows with the amount of training data, although this relation weakens without GoEmotions itself. On GoEmotions, however, the advantage may equally come from its idiosyncratic labelling of \emph{joy} (Section~\ref{sec:disc-datasets}); with a single large dataset the two explanations cannot be separated, and a learning curve on the same data would be needed to tell them apart.

The average gap we observe, about one point although gaps of 4--5 points cannot be excluded, is smaller than the gaps reported by earlier comparisons of prompted LLMs with fine-tuned models on emotions \citep{zhang-etal-2024-sentiment,kocon2023chatgpt,bucher2024fine,meinert2026llms}. Those studies used mostly earlier LLMs, a single prompt and often finer-grained labels, whereas our four coarse classes and an average over ten datasets may favour zero-shot systems; on individual datasets, large differences in both directions remain (Section~\ref{sec:res-h1}).

The reuse of labelled data from another domain behaves differently for the two encoders. After fine-tuning on a different emotion dataset, the public XLM-EMO checkpoint scored on average below the checkpoint itself used without any training (0.596 against 0.616 macro-F1; paired difference $-0.020$, 95\% CI $[-0.031, -0.009]$, a complementary test without multiplicity correction). For tabularisai the opposite holds: fine-tuning on another dataset improves on the public checkpoint (Section~\ref{sec:res-h2}). In our data, out-of-domain fine-tuning therefore lowered the checkpoint trained on real emotion data and improved the one trained on synthetic data, a pattern observed on one checkpoint of each kind. For XLM-EMO, labelled data from a different source may have erased part of what the model had learned from broader emotion data, in line with the observation that fine-tuning can distort pre-trained features and degrade out-of-distribution performance \citep{kumar2022finetuning}. For a practitioner who has labelled data only from another domain, a zero-shot system, or a public checkpoint trained on real emotion data, may be a better choice than fine-tuning on those data.

On the second question, the decision format does not add quality. OpenAI Decisions matches GPT-6 Luna, the LLM on which it runs, and giving Qwen3.5-2B the same output through label scoring leaves its quality unchanged. What the format adds is speed and, for closed models whose API returns no probability for the alternatives, a probability for every option. OpenAI Decisions is about five times faster than GPT-6 Luna, and for the open model label scoring is already faster than generation. The only matched comparison of calibration favours the trained decision model: Strands Decider 2B is better calibrated than label-scored Qwen3.5-2B, built on the same base model (expected calibration error 0.088 against 0.104, Brier score 0.301 against 0.403, both differences significant in the complementary bootstrap; Section~\ref{sec:res-calibration}), but not better than the twice-larger Qwen3.5-4B. Against the fine-tuned encoders, the advantage in calibration disappears in-domain once both are recalibrated, but not out of domain (Section~\ref{sec:res-calibration}). Cost does not separate the systems, at about USD 0.02 per thousand requests for each API. 

Our finding that the decision format costs no quality differs from Emo-Jev \citep{emojev}, where Jev scores below the strongest of five LLMs (Section~\ref{sec:related}). We consider the different composition of the benchmarks the most likely cause: the Emo-Jev average also covers sentiment, sarcasm and humour, which we do not evaluate, whereas on four-class emotion recognition Jev and GPT-6 Luna are equivalent in our data; different LLMs and prompts, and the single runs of that study, may also contribute. 

The comparison between Strands Decider 2B and Qwen3.5-2B shows that a specialised decision model can be strong at small scale. At two billion parameters, the decision model is almost ten points of macro-F1 better than a general LLM of the same size and far less sensitive to the prompt. It is also faster (62 against 143\,ms with English instructions, on the same L4 GPU), but label scoring already brings the general model to 84\,ms in the same conditions, so much of this speed comes from not generating text: at this size the format gives speed, and the training gives quality. Both models start from the same base model, and label scoring shows that giving the general model the decision format without training changes neither its quality nor its instability: the gap reflects how the model was specialised, through the decision adapter and its partial distillation from a larger model, rather than the decision format. Qwen3.5-4B, the distillation teacher, shows how large the gain is: the 2B decision model is equivalent to its teacher, used generatively or with label scoring, so its training transfers the quality of a model twice its size without exceeding it, at a lower latency and with greater stability. Whether an adapter trained without distillation would keep this advantage remains open.

\subsection{Why results differ across datasets}
\label{sec:disc-datasets}

The very large heterogeneity between datasets is the most important qualification of our results. We can describe the two extreme cases but not fully explain them. On GoEmotions, the zero-shot systems err mostly on texts labelled \emph{joy}: Jev assigns 306 of the 2{,}034 such texts to \emph{anger} and 169 to \emph{sadness}, compared with 126 and 40 for the in-domain encoder, and the F1 of \emph{joy} on GoEmotions is 0.81--0.83 for the three leading systems against 0.93 for both in-domain encoders (Table~\ref{tab:per-class}). This pattern is compatible with a label convention that a zero-shot system cannot learn: in GoEmotions, \emph{joy} results from the mapping of many fine-grained positive emotions onto a single class, and the original annotation may not match the everyday meaning of the word. Our data are consistent with this explanation but do not prove it. Strands Decider 2B is an exception: on GoEmotions it scores 0.651, above the three leading systems, with an F1 of 0.87 for \emph{joy}. Its read-out head was trained on undocumented data, so we cannot tell whether it learned a convention closer to that of GoEmotions or saw similar data during training. On BRIGHTER, the opposite happens: the fine-tuned encoder had only 1{,}388 training texts and makes more errors on every class, especially \emph{fear} (Table~\ref{tab:per-class}), whereas the zero-shot systems are not affected by the size of the training split. The size of the training split cannot be the whole explanation, however: EMit and MultiEmotions-It have training splits of similar size (1{,}432 and 1{,}305 texts) and show no significant advantage of the zero-shot systems. Other properties of BRIGHTER, such as its source and its annotation procedure, may contribute, and our design cannot separate them. These two cases are consistent with a general trade-off, which our data suggest but do not establish: fine-tuning captures the conventions of a specific dataset, which helps when the dataset is large and its conventions are idiosyncratic, while zero-shot systems apply a general notion of each emotion, which can help when training data are scarce.

\subsection{Robustness and evaluation practice}
\label{sec:disc-practice}

Robustness to the prompt is a practical requirement, because in a zero-shot setting there is no labelled data on which to choose a formulation. Jev, OpenAI Decisions and GPT-6 Luna satisfy this requirement, and Strands Decider 2B nearly does: the standard deviation of their macro-F1 across four formulations and option orders is below two points on average, although single test sets can vary more, and their preference for any option position is negligible. The small LLM does not, and its preference of 10.6 percentage points of answers for the first option shows that a multiple-choice evaluation of such a model with a single option order can be biased. This confirms, in a multilingual emotion task, the concerns raised by previous work on prompt sensitivity \citep{sclar2024quantifying,mizrahi-etal-2024-state} and on position bias \citep{zheng2024large,pezeshkpour-hruschka-2024-large}, and it supports evaluation over several prompt variants with balanced option orders.

The language of the instructions matters in the same way. With Italian instructions on German, Spanish and French texts, Qwen3.5-2B loses about six points, and English instructions do not help it: its quality does not change and its preference for the first option grows from 10.6 to 17.6 percentage points of answers. Earlier work found English task descriptions better than descriptions in the language of the text for ChatGPT \citep{lai-etal-2023-chatgpt}; for the small open LLM we find the opposite, and for the three leading systems the language changes macro-F1 by less than one point, equivalent to Italian within two points. A practitioner using a small general LLM should therefore write the instructions in the language of the texts, whereas the leading systems can be given the same instructions for all languages.

Our analysis also illustrates how much the conclusions of a comparison depend on the way it is reported. A mean over datasets can hide differences of $\pm0.2$ on individual datasets; the simple mean and the mean weighted by dataset size can disagree on whether a difference exists at all; and the absence of a significant difference is far from a demonstration of equivalence. In our case, the statement that zero-shot systems ``match'' in-domain fine-tuning would have been natural on the basis of the means alone, but it is not supported by the equivalence test and is contradicted on several individual datasets. We therefore recommend that comparisons of this kind report per-dataset paired differences, an explicit equivalence margin with the corresponding test, and a prediction interval for new datasets.

\subsection{Limitations}
\label{sec:limitations}

Several limitations concern the task and the data. We consider only four classes and only texts with a single label, which reduces SemEval-2018 by about 30\% in English and 18\% in Spanish; GoEmotions accounts for 76\% of the English texts; \emph{fear} is almost absent from Universal Joy (8 and 22 texts overall), and without it the macro-F1 of all systems there rises by 0.10--0.19 (Table~\ref{tab:per-class}). Excluding these test sets changes no outcome at the two-point margin (Section~\ref{sec:res-h1}). The smallest test sets (163 and 256 texts) give wide confidence intervals. German combines Reddit and Facebook texts and French has a single native source, so differences between languages cannot be separated from differences between sources. We could not check whether test texts of one dataset appear in the training data of another, and the Hindi training data of XLM-EMO include translations of GoEmotions texts.

Other limitations concern the zero-shot systems. The instructions are in Italian for texts in five languages; their language matters for the small general LLM but changes the leading systems by less than one point, equivalent within two points (Section~\ref{sec:res-h4}), English instructions were tested only on the 2B models, and the translations were checked by one of the authors but not reviewed by native speakers. Each prompt variant changes both the wording and the order of the options, so the two effects cannot be separated, and four variants are a small sample of possible formulations. Jev and GPT-6 Luna are not deterministic: on the 1{,}431 duplicated BRIGHTER texts, their predictions for the two copies of a text differ in 1.1\% and 4.0\% of the 5{,}724 copy pairs, against 0\% for all other systems; a second run shows that their small variant effects are partly due to the prompt and partly to run-to-run variation (Section~\ref{sec:res-h4}). Absolute latencies depend on the network, on the day and on the GPU, and the prices of two of the three APIs could not be verified (Section~\ref{sec:res-h3}). OpenAI Decisions was run with its default settings, so its latency advantage over GPT-6 Luna may partly reflect a different amount of internal computation, which we cannot verify. The closed systems were called through aliases and endpoints that can change over time (\texttt{jev-latest}, which served version 1.13.0, and the OpenAI APIs, which return no dated version), between 19 October and 2 November 2026: the released predictions allow every number to be recomputed, but re-running the experiments may give different results, and contamination with the public test sets cannot be ruled out. Label scoring separates format from training for the 2B pair, but not the distillation from the adapter, and it could not be applied to GPT-6 Luna, whose API returns the probability of the chosen answer only.

The fine-tuned references also have limitations. Both starting checkpoints were already trained on emotion data, and the eleven classes of tabularisai were collapsed to our four. Each model uses a single source dataset, whereas a model trained on several sources might generalise better. The out-of-domain condition includes the same corpus in another language, which favours the encoders. Three seeds give only a rough estimate of the variability due to the seed. On the statistical side, the margin of two points was fixed in advance but checked against seed variability afterwards; with 8--10 datasets the Wilcoxon test and the random-effects model have little power, and H1 could not reach equivalence with test sets of this size (Section~\ref{sec:stats}); the hypotheses and the analysis plan were written after the mean macro-F1 of every system was known and were not preregistered, the verdict rules for H3 and H4 were fixed afterwards, the three leading zero-shot systems were identified on the descriptive results, and the complementary analyses were added after the first results (Appendix~\ref{app:deviations}); the recalibration of the encoders and decision models estimates the temperature on halves of the test sets, because predictions on the validation splits were not stored; out of domain, this uses labelled texts of the target domain, which that scenario lacks; and removing the class weights corrects the output of the trained models, which is not the same as training them without weights.

Finally, all datasets are public research datasets of social-media texts, used under their licences, and no new data were collected; the tabularisai checkpoint is released under a non-commercial licence (CC BY-NC 4.0). Emotion classifiers can be misused to profile individuals. Our results concern aggregate accuracy on public benchmarks and do not support decisions about individuals.

\section{Conclusions}
\label{sec:conclusions}

We presented the first comparison of several decision models, systems that choose among given options instead of generating text, with general LLMs and fine-tuned encoders, on four-class emotion classification in 10 datasets and five languages, with paired tests of difference and of equivalence. Used without any training, the three leading zero-shot systems, Jev, OpenAI Decisions and GPT-6 Luna, come on average within about one point of an encoder fine-tuned on the target dataset, but our test sets cannot exclude a gap of 4--5 points, and on a new dataset either approach could come out ahead. They outperform encoders fine-tuned on a different dataset, by 15 points against tabularisai and, on our datasets only, by 8 points against XLM-EMO, and, in an exploratory comparison, the strongest public emotion model used as it is by about six points.

The decision format does not add quality. OpenAI Decisions matches GPT-6 Luna, the LLM on which it runs, at one fifth of its median latency, and the ten-point advantage of Strands Decider 2B over Qwen3.5-2B comes from its training, as label scoring of the general model shows; Strands Decider 2B matches Qwen3.5-4B, the twice-larger model from which it was partly distilled. What the format adds is speed, a probability for every option where the API of a general LLM returns none, and, in the one matched comparison with a model of the same size, better calibrated probabilities; cost does not separate the API-based systems. The three leading systems are robust to the wording of the question, to the order of the options and to instructions in the language of the text, which change their macro-F1 by less than one point; only the small general LLM is sensitive to them.

These findings suggest a simple rule of thumb. Without labelled data from the target domain, the leading zero-shot systems are a strong and robust default, and a decision model adds speed and per-option probabilities, which are better calibrated than those of an encoder fine-tuned on another dataset when no labelled data are available to recalibrate it; with a small general LLM, the instructions should be written in the language of the texts. With abundant labelled data, fine-tuning may be preferable: it won clearly on GoEmotions, the one large dataset we studied, although we cannot tell whether because of its size or of its labelling conventions. Labelled data from another domain can even harm a model already trained on real emotion data, as it did for XLM-EMO.

Beyond these systems, our analysis shows how easily a mean over a few datasets can be over-read. We recommend that comparisons of this kind report per-dataset paired differences, an explicit equivalence margin with the corresponding test and a prediction interval for new datasets; that they evaluate each system over several prompt variants with balanced option orders; and that a decision model be compared with a general model built on the same base and used through label scoring, to separate the effect of the format from that of training. Future work should measure how many labelled examples a fine-tuned encoder needs to overtake a decision model, train a decision adapter without distillation to isolate its contribution, and extend the comparison to more classes and to texts with several emotions.

\paperbibliography

\appendix

\section{Dataset preparation}
\label{app:data}

\paragraph{Excluded resources.} Some candidate resources were excluded on purpose: the French and German portions of the XLM-EMO training data, which are machine translations of Spanish tweets rather than native texts; EmoDIFT, a French corpus that has no \emph{joy} class; and the French-emotion dataset on Hugging Face, which was generated synthetically by an LLM.

\paragraph{EmoEvent split.} The cleaned release of EmoEvent has numeric identifiers, so its official split was reconstructed by matching the texts against the official split files. Texts that could not be matched unambiguously (160 in English and 107 in Spanish) were assigned to no split.

\paragraph{Duplicates in BRIGHTER.} The BRIGHTER test split, built from tracks A and C of SemEval-2025 Task~11 \citep{muhammad-etal-2025-semeval}, contains each of its 1{,}431 texts twice. Group resampling already accounts for this duplication in the confidence intervals, and recomputing all results with a single copy of each text changes no mean difference by more than $10^{-4}$.

\section{Deviations from the analysis plan}
\label{app:deviations}

The analysis followed a plan written before any paired test was computed. Four changes were made before the paired tests. First, the XLM-EMO matrix starts from the public \texttt{xlm-emo-t} checkpoint rather than from XLM-T base, because it is the only XLM-EMO run for which per-text predictions were stored. Second, invalid answers count as errors for all systems; re-querying the 9 invalid answers of GPT-6 Luna changes its mean score by less than $10^{-4}$. Third, the H2 family includes the in-domain minus out-of-domain loss of each encoder. Fourth, the exploratory comparison with the public checkpoints used without training was added.

The following analyses were added after the first results and are reported as complementary: the random-effects meta-analysis for all pairs; for H1, the per-dataset equivalence tests, the alternative aggregations, the training-size moderator, the analysis of BRIGHTER without duplicates, the analysis without Universal Joy, the smallest equivalence margin $\delta_{\min}$ of each pair and the precision calculation; the inclusion of Qwen3.5-2B in the H1 family, which makes the Holm correction stricter for all pairs of the family (it was not added to the H2 family, because the plan included it only as the size-matched control for H3); the three tests of H4, which was descriptive in the plan, and the verdict rules of H3 and H4; and, for H2, the split of the sources by language, the comparison with the best source and the paired comparison of each out-of-domain encoder with its public checkpoint (Section~\ref{sec:disc-when}); and the recalibration of encoders and decision models, with temperature scaling and with the removal of the class weights of the loss, its sensitivity analysis and the paired bootstrap of the calibration differences, whose comparisons were fixed in writing after the point estimates were known (Section~\ref{sec:res-calibration}). Five sets of runs were added between 28 October and 2 November 2026, after a review of the first version: label scoring of Qwen3.5-2B and Qwen3.5-4B and generative Qwen3.5-4B, the 2B models with instructions in English and in the language of the text, the three leading systems and Qwen3.5-4B with instructions in the language of the text, a controlled latency measurement and a second run of Jev and GPT-6 Luna (Section~\ref{sec:complementary}). They are complementary and do not change any verdict.

The plan numbered the hypotheses differently: H2 of this paper (out-of-domain) was H4 in the plan, H3 (decision models versus LLMs) was H2 and H4 (prompt sensitivity) was H3; the content of the hypotheses is unchanged. The plan was written after the descriptive macro-F1 of every system was known but before any paired test, and it fixed the pairs of each family; the three leading zero-shot systems were therefore identified on descriptive results.

A fifth hypothesis of the original plan, according to which an emotion-oriented starting checkpoint fine-tunes better than XLM-T base, was dropped, because the run that started from XLM-T base has no stored per-text predictions. The comparison between the two encoders is reported as exploratory (Table~\ref{tab:exploratory}).

\ifdefined\mdpiversion\else
\section{Statistical details}
\label{app:statdetails}
\input{appendix_stats}
\fi

\section{Prompt variants}
\label{app:prompts}

Table~\ref{tab:prompts} lists the four question variants and the order in which the options were presented in each of them, and Table~\ref{tab:classdesc} the descriptions of the four classes, identical in all variants. The option keys (\texttt{joy}, \texttt{sadness}, \texttt{anger}, \texttt{fear}) are in English in every condition.

\paragraph{What each system receives.} Jev and OpenAI Decisions receive the question, the four options with their descriptions in the order of the variant, and the text preceded by the label \emph{Testo} (``Text''). GPT-6 Luna and Qwen3.5-2B receive a single prompt: the question; the sentence \emph{Scegli una sola classe tra le seguenti:} (``Choose exactly one of the following classes:''); one line ``\texttt{- class: description}'' per option, in the order of the variant; the sentence \emph{Rispondi solo con il nome della classe (joy, sadness, anger oppure fear), senza altre parole.} (``Answer only with the class name (joy, sadness, anger or fear), no other words.''), with the classes in the order of the variant; and the text after \emph{Testo:}. Label scoring uses the same prompt. Strands Decider 2B receives the question, the descriptions and the text with its label, and builds the rest of its prompt internally.

\paragraph{Translations.} For the complementary runs, the question, the class descriptions and the sentences above were translated into English, German, Spanish and French with Claude Sonnet 5.5 and checked by one of the authors; they were not reviewed by native speakers. The texts in all five languages are released with the code; Tables~\ref{tab:prompts} and~\ref{tab:classdesc} give the English version.

\begin{table}[h]
\centering
\small
\setlength{\tabcolsep}{3pt}
\begin{tabular}{cp{0.55\linewidth}l}
\toprule
\# & Question (Italian; English version) & Option order \\
\midrule
1 & \textit{Quale emozione principale esprime questo messaggio?} (Which main emotion does this message express?) & joy, sad., ang., fear \\
2 & \textit{Qual è lo stato emotivo dominante di chi ha scritto questo testo?} (What is the dominant emotional state of the person who wrote this text?) & sad., fear, joy, ang. \\
3 & \textit{Con quale emozione si può descrivere meglio questo commento?} (Which emotion best describes this comment?) & ang., joy, fear, sad. \\
4 & \textit{Quale emozione prevale in questo testo?} (Which emotion prevails in this text?) & fear, ang., sad., joy \\
\bottomrule
\end{tabular}
\caption{The four question variants. The option orders form a Latin square.}
\label{tab:prompts}
\end{table}

\begin{table}[h]
\centering
\small
\setlength{\tabcolsep}{3pt}
\begin{tabular}{lp{0.37\linewidth}p{0.37\linewidth}}
\toprule
Class & Description (Italian, as sent) & English version \\
\midrule
joy & gioia, felicita, entusiasmo, divertimento, affetto, apprezzamento & joy, happiness, enthusiasm, amusement, affection, appreciation \\
sadness & tristezza, dolore, lutto, delusione, malinconia, compassione & sadness, grief, mourning, disappointment, melancholy, compassion \\
anger & rabbia, indignazione, irritazione, protesta, sdegno & anger, indignation, irritation, protest, outrage \\
fear & paura, ansia, preoccupazione, timore, allarme & fear, anxiety, worry, apprehension, alarm \\
\bottomrule
\end{tabular}
\caption{Descriptions of the four classes, identical in all variants (``felicita'' without accent, as in the prompts sent to the systems).}
\label{tab:classdesc}
\end{table}

\ifdefined\mdpiversion\else
\section{Additional results}
\label{app:extra}

This appendix collects the per-dataset results of H1 (Table~\ref{tab:h1-per-dataset}), the exploratory training-size analysis (Table~\ref{tab:moderator}), the alternative aggregations of H1 (Table~\ref{tab:aggregation}), H1, H2 and H3 without Universal Joy (Table~\ref{tab:no-uj}), the Wilcoxon tests of all comparisons (Table~\ref{tab:wilcoxon}), the complementary analyses of H2 (Table~\ref{tab:h4-extra}), the exploratory comparisons (Table~\ref{tab:exploratory}), the F1 of each class (Table~\ref{tab:per-class}), the complementary runs (Tables~\ref{tab:complementary}, \ref{tab:complementary-h4} and~\ref{tab:controlled-latency}), the bootstrap of the calibration differences (Table~\ref{tab:calibration-tests}) and the position bias of H4 (Figure~\ref{fig:position}).

\begin{figure*}[!htb]
\centering
\includegraphics[width=\textwidth]{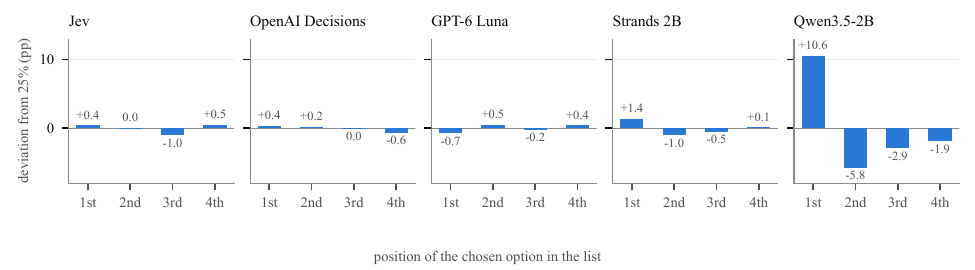}
\caption{H4: deviation of the share of answers given to the option in each list position from the 25\% expected under the Latin-square design (percentage points).}
\label{fig:position}
\end{figure*}

\input{tables/h1_per_dataset}
\ifdefined\mdpiversion\else
\input{tables/h1_moderator}
\input{tables/h1_aggregation}
\input{tables/h1_no_uj}
\input{tables/wilcoxon}
\input{tables/h4_extra}
\input{tables/exploratory}
\input{tables/per_class}
\input{tables/complementary}
\input{tables/complementary_h4}
\input{tables/controlled_latency}
\input{tables/calibration_tests}
\fi
\fi

\section{Compute and reproducibility}
\label{app:compute}

All systems were run on AWS SageMaker. The calls to the three API-based systems were made from CPU instances, Strands Decider 2B and Qwen3.5-2B ran on single GPUs one request at a time, and the fine-tuned encoders were evaluated on T4 GPUs in batches of 32 texts. The zero-shot suite (5 systems, 4 variants, 12{,}633 texts) was run on 19 October 2026 and took 3\,h\,18\,min of wall-clock time with the systems running in parallel and cost USD 7.98 of SageMaker compute, to which the API fees, billed separately, must be added. Fine-tuning 30 tabularisai models cost USD 10.22 and fine-tuning 24 XLM-EMO models USD 6.37, with the models trained in series on single GPUs (L4, L40S or A10G). The open zero-shot models were used at the following Hugging Face commits: \texttt{15852e8c1636} for Qwen3.5-2B, \texttt{851bf6e806ef} for Qwen3.5-4B and \texttt{2b52a6235c1b} for Strands Decider 2B. Once trained, the encoders classify a text in a median of 2.5--3.3\,ms on a T4 GPU in batches of 32. Table~\ref{tab:latency}\ifdefined\mdpiversion{} in the Supplementary Materials\fi{} gives the median, mean and 95th percentile of the latency of every system.

\ifdefined\mdpiversion\else\input{tables/latency}\fi
 The complementary runs of 28 October to 2 November 2026 cost about USD 20 of SageMaker compute for the three local models and an estimated USD 3.5 of API fees. All statistics were computed locally from the stored predictions. The bootstrap seeds are 20261022 plus the alphabetical index of the test set. The per-text predictions of every system and the statistical scripts will be released in a public repository upon publication of this article.

%% file: tables/overview.tex
\begin{table*}[t]
\centering
\small
\setlength{\tabcolsep}{3pt}
\begin{adjustbox}{max width=\linewidth}
\begin{tabular}{lcccccccccccc}
\toprule
System & BRI\,de & EMit\,it & EmE\,en & EmE\,es & GoE\,en & MEm\,it & SE18\,en & SE18\,es & UJ\,de & UJ\,fr & Mean\textsubscript{10} & Mean\textsubscript{8} \\
\midrule
\multicolumn{13}{l}{\textit{Zero-shot: decision models and LLMs (mean of 4 prompt variants)}} \\
Jev & 0.845 & 0.765 & 0.783 & 0.817 & 0.598 & 0.639 & 0.798 & 0.847 & 0.486 & 0.502 & 0.708 & 0.679 \\
OpenAI Decisions & 0.839 & 0.794 & 0.814 & 0.808 & 0.573 & 0.583 & 0.792 & 0.828 & 0.504 & 0.534 & 0.707 & 0.681 \\
GPT-6 Luna & 0.854 & 0.821 & 0.789 & 0.796 & 0.563 & 0.589 & 0.784 & 0.838 & 0.502 & 0.529 & 0.706 & 0.680 \\
Strands 2B & 0.757 & 0.694 & 0.750 & 0.793 & 0.651 & 0.491 & 0.817 & 0.805 & 0.441 & 0.486 & 0.669 & 0.633 \\
Qwen3.5-2B & 0.480 & 0.671 & 0.629 & 0.648 & 0.588 & 0.471 & 0.722 & 0.727 & 0.362 & 0.417 & 0.571 & 0.533 \\
\midrule
\multicolumn{13}{l}{\textit{Fine-tuned, in-domain (mean of 3 seeds)}} \\
tabularisai\textsubscript{in} & 0.712 & 0.809 & 0.764 & 0.820 & 0.778 & 0.575 & 0.797 & 0.816 & 0.512 & 0.583 & 0.717 & 0.694 \\
XLM-EMO\textsubscript{in} & 0.696 & 0.791 & 0.779 & 0.817 & 0.770 & 0.568 & -- & -- & 0.552 & 0.549 & -- & 0.690 \\
\midrule
\multicolumn{13}{l}{\textit{Fine-tuned, out-of-domain (mean over source datasets and seeds)}} \\
tabularisai\textsubscript{out} & 0.560 & 0.611 & 0.623 & 0.678 & 0.517 & 0.478 & 0.632 & 0.610 & 0.408 & 0.417 & 0.553 & 0.536 \\
XLM-EMO\textsubscript{out} & 0.588 & 0.725 & 0.707 & 0.760 & 0.606 & 0.496 & -- & -- & 0.431 & 0.454 & -- & 0.596 \\
\midrule
\multicolumn{13}{l}{\textit{Public encoders, no training on our data}} \\
tabularisai\textsubscript{zs} & 0.562 & 0.652 & 0.580 & 0.631 & 0.401 & 0.474 & 0.606 & 0.633 & 0.346 & 0.367 & 0.525 & 0.502 \\
XLM-EMO\textsubscript{zs} & 0.620 & 0.731 & 0.720 & 0.768 & 0.659 & 0.537 & -- & -- & 0.438 & 0.452 & -- & 0.616 \\
\bottomrule
\end{tabular}
\end{adjustbox}
\caption{Macro-F1 on the test split of each dataset. Mean\textsubscript{10}: unweighted mean over the 10 test sets; Mean\textsubscript{8}: over the 8 test sets without SemEval-2018 (XLM-EMO saw SemEval-2018 during pre-training). Group-bootstrap 95\% CIs for every cell are in the released files.}
\label{tab:overview}
\end{table*}

%% file: tables/verdicts_numeric.tex
\begin{table*}[t]
\centering
\small
\setlength{\tabcolsep}{5pt}
\begin{adjustbox}{max width=\linewidth}
\begin{tabular}{lcccc}
\toprule
System & \makecell{H1: vs.\ in-domain\\$\Delta$ [90\% CI]} & \makecell{H2b: vs.\ out-of-domain\\tabularisai / XLM-EMO} & \makecell{H3: vs.\ LLM\\$\Delta$; median latency} & \makecell{H4: prompt\\SD (UB\textsubscript{90})} \\
\midrule
Jev & \makecell{\textbf{inconclusive}\\$-$0.9 [$-$4.0, +0.7]} & \makecell{\textbf{supported}\\+15.5$^{\dagger}$ / +8.3} & \makecell{\textbf{unmatched$^{a}$}\\+0.1; 220 vs 1{,}107\,ms} & \makecell{\textbf{supported}\\0.8 (1.26)} \\
OpenAI Decisions & \makecell{\textbf{inconclusive}\\$-$1.0 [$-$3.9, +0.6]} & \makecell{\textbf{supported}\\+15.4$^{\dagger}$ / +8.6} & \makecell{\textbf{supported$^{b}$}\\+0.1; 229 vs 1{,}107\,ms} & \makecell{\textbf{supported}\\0.9 (1.51)} \\
GPT-6 Luna & \makecell{\textbf{inconclusive}\\$-$1.0 [$-$4.0, +0.5]} & \makecell{\textbf{supported}\\+15.3$^{\dagger}$ / +8.5} & -- & \makecell{\textbf{supported}\\1.3 (1.77)} \\
Strands 2B & \makecell{\textbf{worse}\\$-$4.8 [$-$7.5, $-$3.6]} & \makecell{\textbf{supported}\\+11.5$^{\dagger}$ / +3.7} & \makecell{\textbf{supported}\\+9.7; 65 vs 146\,ms} & \makecell{\textbf{inconclusive}\\1.2 (2.04)} \\
Qwen3.5-2B & \makecell{\textbf{worse}\\$-$14.5 [$-$16.3, $-$13.2]} & -- & -- & \makecell{\textbf{rejected}\\7.3 (8.50)} \\
\bottomrule
\end{tabular}
\end{adjustbox}
\caption{Summary of the results, in macro-F1 points, with the outcome of each hypothesis in bold (rules of Section~\ref{sec:stats}). H1: mean difference from tabularisai\textsubscript{in} with its 90\% CI. H2b: mean difference from the out-of-domain encoders; $^{\dagger}$also significant under the random-effects model; H2a is supported for both encoders. H3: difference from the general LLM of the pair and median latencies of the main runs. H4: mean SD across the four prompt variants, with the upper bound of its 90\% CI (\emph{stable} if below 2). $^{a}$Unmatched pair: same quality and lower latency, but no verdict on H3, since the architecture of Jev is not disclosed. $^{b}$Latency ratio confirmed by a controlled sequential measurement. --: not tested.}
\label{tab:verdicts}
\end{table*}

%% file: tables/h1.tex
\begin{table*}[t]
\centering
\small
\setlength{\tabcolsep}{4pt}
\begin{adjustbox}{max width=\linewidth}
\begin{tabular}{llccccclccccc}
\toprule
 & & \multicolumn{8}{c}{Mean over test sets (primary analysis)} & \multicolumn{3}{c}{Random effects} \\
\cmidrule(lr){3-10}\cmidrule(lr){11-13}
Zero-shot & Reference & $n$ & $\Delta$ & 95\% CI & 90\% CI & $p_{\text{Holm}}$ & TOST ($\delta{=}0.02$) & $\delta_{\min}$ & $+/=/-$ & $\Delta_{\text{RE}}$ & 95\% CI & 95\% PI \\
\midrule
Jev & tabularisai\textsubscript{in} & 10 & $-$0.009 & [$-$0.045, 0.010] & [$-$0.040, 0.007] & 1.000 & inconclusive & 0.040 & 1/7/2 & $-$0.011 & [$-$0.074, 0.052] & [$-$0.22, 0.20] \\
OpenAI Decisions & tabularisai\textsubscript{in} & 10 & $-$0.010 & [$-$0.044, 0.009] & [$-$0.039, 0.006] & 1.000 & inconclusive & 0.039 & 1/8/1 & $-$0.010 & [$-$0.074, 0.053] & [$-$0.22, 0.20] \\
GPT-6 Luna & tabularisai\textsubscript{in} & 10 & $-$0.010 & [$-$0.045, 0.008] & [$-$0.040, 0.005] & 1.000 & inconclusive & 0.040 & 1/8/1 & $-$0.011 & [$-$0.078, 0.056] & [$-$0.24, 0.21] \\
Strands 2B & tabularisai\textsubscript{in} & 10 & $-$0.048 & [$-$0.079, $-$0.033] & [$-$0.075, $-$0.036] & 0.002 & different & 0.075 & 0/7/3 & $-$0.041 & [$-$0.085, 0.003] & [$-$0.18, 0.10] \\
Qwen3.5-2B & tabularisai\textsubscript{in} & 10 & $-$0.145 & [$-$0.166, $-$0.129] & [$-$0.163, $-$0.132] & 0.002 & different & 0.163 & 0/1/9 & $-$0.147 & [$-$0.184, $-$0.111] & [$-$0.27, $-$0.03] \\
\midrule
Jev & XLM-EMO\textsubscript{in} & 8 & $-$0.011 & [$-$0.055, 0.012] & [$-$0.049, 0.008] & 1.000 & inconclusive & 0.049 & 1/6/1 & $-$0.014 & [$-$0.097, 0.068] & [$-$0.26, 0.24] \\
OpenAI Decisions & XLM-EMO\textsubscript{in} & 8 & $-$0.009 & [$-$0.051, 0.014] & [$-$0.045, 0.010] & 1.000 & inconclusive & 0.045 & 1/6/1 & $-$0.009 & [$-$0.093, 0.075] & [$-$0.27, 0.25] \\
GPT-6 Luna & XLM-EMO\textsubscript{in} & 8 & $-$0.010 & [$-$0.051, 0.013] & [$-$0.046, 0.009] & 1.000 & inconclusive & 0.046 & 1/6/1 & $-$0.011 & [$-$0.100, 0.079] & [$-$0.29, 0.26] \\
Strands 2B & XLM-EMO\textsubscript{in} & 8 & $-$0.058 & [$-$0.095, $-$0.038] & [$-$0.090, $-$0.042] & 0.002 & different & 0.090 & 1/5/2 & $-$0.050 & [$-$0.103, 0.002] & [$-$0.20, 0.10] \\
Qwen3.5-2B & XLM-EMO\textsubscript{in} & 8 & $-$0.157 & [$-$0.183, $-$0.135] & [$-$0.179, $-$0.139] & 0.002 & different & 0.179 & 0/2/6 & $-$0.169 & [$-$0.199, $-$0.139] & [$-$0.24, $-$0.10] \\
\bottomrule
\end{tabular}
\end{adjustbox}
\caption{H1: zero-shot systems minus fine-tuned encoders trained on the test dataset (in-domain). $\Delta$: mean macro-F1 difference over test sets with group-bootstrap 95\% and 90\% CIs; $p_{\text{Holm}}$: bootstrap $p$, Holm-adjusted within the family. TOST: \emph{equivalent} if the 90\% CI lies within $\pm0.02$, \emph{different} if $p_{\text{Holm}}<0.05$, otherwise \emph{inconclusive}. $\delta_{\min}$: smallest equivalence margin, the larger absolute bound of the 90\% CI (the TOST concludes equivalence for any larger margin). $+/=/-$: test sets where the zero-shot system is significantly better / not different / worse (per-dataset bootstrap, Holm). Random effects: REML with Hartung--Knapp CI; PI: 95\% prediction interval for a new dataset.}
\label{tab:h1}
\end{table*}

%% file: tables/h4.tex
\begin{table*}[t]
\centering
\small
\setlength{\tabcolsep}{4pt}
\begin{adjustbox}{max width=\linewidth}
\begin{tabular}{llccccccc}
\toprule
A & B & $n$ & $\Delta$ & 95\% CI & $p_{\text{Holm}}$ & $+/=/-$ & RE $p_{\text{Holm}}$ & RE 95\% PI \\
\midrule
\multicolumn{9}{l}{\textit{Loss of the fine-tuned encoder outside its training dataset}} \\
tabularisai\textsubscript{in} & tabularisai\textsubscript{out} & 10 & +0.163 & [0.150, 0.191] & 0.002 & 9/1/0 & $<$0.001 & [0.07, 0.27] \\
XLM-EMO\textsubscript{in} & XLM-EMO\textsubscript{out} & 8 & +0.095 & [0.078, 0.131] & 0.002 & 6/2/0 & 0.001 & [0.001, 0.19] \\
\midrule
\multicolumn{9}{l}{\textit{Zero-shot against the out-of-domain fine-tuned encoder}} \\
Jev & tabularisai\textsubscript{out} & 10 & +0.155 & [0.137, 0.169] & 0.002 & 8/2/0 & $<$0.001 & [$-$0.01, 0.32] \\
OpenAI Decisions & tabularisai\textsubscript{out} & 10 & +0.154 & [0.138, 0.168] & 0.002 & 9/1/0 & $<$0.001 & [$-$0.004, 0.31] \\
GPT-6 Luna & tabularisai\textsubscript{out} & 10 & +0.153 & [0.137, 0.167] & 0.002 & 9/1/0 & $<$0.001 & [$-$0.02, 0.33] \\
Strands 2B & tabularisai\textsubscript{out} & 10 & +0.115 & [0.103, 0.126] & 0.002 & 7/3/0 & 0.002 & [$-$0.04, 0.27] \\
Jev & XLM-EMO\textsubscript{out} & 8 & +0.083 & [0.063, 0.102] & 0.002 & 1/7/0 & 0.085 & [$-$0.13, 0.29] \\
OpenAI Decisions & XLM-EMO\textsubscript{out} & 8 & +0.086 & [0.066, 0.104] & 0.002 & 3/4/1 & 0.085 & [$-$0.12, 0.30] \\
GPT-6 Luna & XLM-EMO\textsubscript{out} & 8 & +0.085 & [0.065, 0.103] & 0.002 & 2/5/1 & 0.085 & [$-$0.14, 0.31] \\
Strands 2B & XLM-EMO\textsubscript{out} & 8 & +0.037 & [0.023, 0.051] & 0.002 & 2/6/0 & 0.112 & [$-$0.11, 0.19] \\
\bottomrule
\end{tabular}
\end{adjustbox}
\caption{H2: out-of-domain comparison. \textsubscript{out}: mean over the encoders fine-tuned on each of the other datasets (9 sources for tabularisai, 7 for XLM-EMO, 3 seeds each). Columns as in Table~\ref{tab:h1}; RE: random-effects (Hartung--Knapp) test of the mean difference; PI: 95\% prediction interval for a new dataset. All $p$ are Holm-adjusted within the family. The Wilcoxon tests are in Table~\ref{tab:wilcoxon}. A bound that would round to zero is shown with one more decimal, to keep its side of zero.}
\label{tab:h4}
\end{table*}

%% file: tables/h2.tex
\begin{table*}[t]
\centering
\small
\setlength{\tabcolsep}{3pt}
\begin{tabular}{lccc}
\toprule
 & OpenAI Dec. & Jev & Strands 2B \\
 & $-$ GPT-6 Luna & $-$ GPT-6 Luna & $-$ Qwen3.5-2B \\
\midrule
$\Delta$ macro-F1 & +0.001 & +0.001 & +0.097 \\
95\% CI & [$-$0.006, 0.008] & [$-$0.009, 0.012] & [0.075, 0.109] \\
90\% CI & [$-$0.005, 0.007] & [$-$0.007, 0.010] & [0.078, 0.106] \\
TOST $\delta{=}0.01$ & equivalent (0.006) & equivalent (0.048) & different (1.000) \\
TOST $\delta{=}0.02$ & equivalent ($<$0.001) & equivalent (0.002) & different (1.000) \\
$+/=/-$ test sets & 0/10/0 & 1/9/0 & 6/4/0 \\
$\Delta_{\text{RE}}$ [95\% CI] & +0.002 [$-$0.01, 0.01] & +0.003 [$-$0.02, 0.02] & +0.100 [0.05, 0.15] \\
\midrule
Median latency (ms) & 229 vs 1107 & 220 vs 1107 & 65 vs 146 \\
\bottomrule
\end{tabular}
\caption{H3: decision models against general-purpose LLMs. Quality: mean macro-F1 difference over the 10 test sets with the tests of \S\ref{sec:stats} (in parentheses, the TOST $p$-value, i.e.\ the larger of the two one-sided $p$-values: a value close to 1, as for Strands 2B, means that the difference clearly exceeds the margin, which is why the outcome is \emph{different}). Latency: median per request over all 50{,}532 requests (4 variants); API latencies include the network, GPU latencies are one request at a time, so the two groups are not comparable with each other.}
\label{tab:h2}
\end{table*}

%% file: tables/h3.tex
\begin{table}[t]
\centering
\small
\setlength{\tabcolsep}{3pt}
\begin{adjustbox}{max width=\linewidth}
\begin{tabular}{lccccc}
\toprule
 & \multicolumn{3}{c}{Across prompt variants} & \multicolumn{2}{c}{Position bias} \\
\cmidrule(lr){2-4}\cmidrule(lr){5-6}
System & SD & UB\textsubscript{90} & $p_{\text{Holm}}$ & max (pp) & $p_{\text{Holm}}$ \\
\midrule
Jev & 0.008 & 0.0126 & 0.030 & $-$1.0 (pos.\,3) & 0.001 \\
OpenAI Decisions & 0.009 & 0.0151 & 0.182 & $-$0.6 (pos.\,4) & 0.001 \\
GPT-6 Luna & 0.013 & 0.0177 & 0.001 & $-$0.7 (pos.\,1) & 0.001 \\
Strands 2B & 0.012 & 0.0204 & 0.182 & +1.4 (pos.\,1) & 0.001 \\
Qwen3.5-2B & 0.073 & 0.0850 & 0.001 & +10.6 (pos.\,1) & 0.001 \\
\bottomrule
\end{tabular}
\end{adjustbox}
\caption{H4: sensitivity to the prompt. SD: standard deviation of macro-F1 across the 4 variants, averaged over the 10 test sets; UB\textsubscript{90}: upper bound of its group-bootstrap 90\% CI (\emph{stable} if $<0.02$); $p_{\text{Holm}}$: permutation test of no variant effect (answers shuffled across variants within each text, 5{,}000 permutations, Holm over the 5 systems). Position bias: largest deviation from the 25\% share expected under the Latin-square design, in percentage points.}
\label{tab:h3}
\end{table}

%% file: tables/calibration.tex
\begin{table}[t]
\centering
\small
\setlength{\tabcolsep}{4pt}
\begin{tabular}{lcccc}
\toprule
 & \multicolumn{2}{c}{As returned} & \multicolumn{2}{c}{Recalibrated} \\
\cmidrule(lr){2-3}\cmidrule(lr){4-5}
System & ECE & Brier & ECE & Brier \\
\midrule
Jev & 0.098 & 0.286 & 0.047 & 0.270 \\
OpenAI Decisions & 0.090 & 0.280 & 0.041 & 0.265 \\
Strands 2B & 0.088 & 0.301 & 0.037 & 0.292 \\
tabularisai\textsubscript{in} & 0.129 & 0.305 & 0.047 & 0.272 \\
XLM-EMO\textsubscript{in} & 0.134 & 0.308 & 0.052 & 0.267 \\
tabularisai\textsubscript{out} & 0.226 & 0.527 & 0.059 & 0.440 \\
XLM-EMO\textsubscript{out} & 0.195 & 0.443 & 0.059 & 0.372 \\
\midrule
\multicolumn{5}{l}{\emph{Label scoring (complementary runs)}} \\
Qwen3.5-2B & 0.104 & 0.403 & 0.058 & 0.379 \\
Qwen3.5-4B & 0.074 & 0.309 & 0.046 & 0.301 \\
\bottomrule
\end{tabular}
\caption{Calibration of the systems that return class probabilities: expected calibration error (15 equal-width bins on the top probability) and multiclass Brier score, averaged over test sets (and over variants or seeds). GPT-6 Luna and Qwen3.5-2B used generatively answer in free text and return no probabilities; the last block gives the two Qwen models with label scoring (summed log-probabilities). Recalibrated (complementary): temperature scaling, with the logits, recovered from the stored probabilities, divided by a calibration temperature estimated on one half of each test set and applied to the other (groups of identical texts kept together; Section~\ref{sec:res-calibration}); the mean calibration temperature is 2.8 for the encoders, about 1.4 for Jev and OpenAI Decisions and 0.8 for Strands 2B, which is slightly underconfident; the predicted class, and hence macro-F1, does not change.}
\label{tab:calibration}
\end{table}

%% file: appendix_stats.tex
\paragraph{Bootstrap.} Within each test set, groups of texts with the same normalised-text key are resampled with replacement, $B=10{,}000$ times, with a fixed seed per test set (20261022 plus the alphabetical index of the test set). When a class is neither present in the gold labels nor predicted in a resample, it is left out of the macro average. In Universal Joy, where \emph{fear} has 1 and 2 test texts, the average therefore covers three or four classes depending on the resample, and a single false \emph{fear} prediction in a resample without gold \emph{fear} sets that F1 to zero; this makes the per-dataset intervals of Universal Joy wide and asymmetric, which is one reason for the analysis without these two test sets (Section~\ref{sec:res-h1}). Confidence intervals are percentile intervals. The two-sided bootstrap $p$-value of an observed difference is $p = 2(c+1)/(B+1)$, where $c$ is the number of resamples in which the difference has the sign opposite to the observed one; its minimum is therefore 0.0002. In a family of 10 pairs the smallest Holm-adjusted value is consequently 0.002, which explains why this value recurs in the tables.

\paragraph{Equivalence.} The TOST procedure rejects both one-sided null hypotheses $\Delta \le -\delta$ and $\Delta \ge +\delta$ at level 0.05 exactly when the 90\% CI lies within $[-\delta, +\delta]$. The one-sided $p$-values reported in the text are the bootstrap proportions $(c+1)/(B+1)$, where $c$ is the number of resamples beyond each margin. As is standard, the TOST uses the unadjusted interval.

\paragraph{Random effects.} The model assumes that each test set has its own true difference; these true differences vary around a common mean with a between-dataset standard deviation $\tau$, and the observed difference on each test set adds sampling error to its true difference. For each pair, $y_j$ is the observed difference on test set $j$ and $v_j$ its bootstrap variance. The between-dataset variance $\tau^2$ is estimated by restricted maximum likelihood \citep{viechtbauer2005bias}. The CI of the mean difference uses the Hartung--Knapp standard error with a $t$ distribution on $K-1$ degrees of freedom, where $K$ is the number of test sets \citep{hartung2001refined,inthout2014hartung}, and the 95\% prediction interval uses a $t$ distribution on $K-2$ degrees of freedom \citep{higgins2009reevaluation}. Heterogeneity is summarised by Cochran's $Q$ and by $I^2$ \citep{higgins2002quantifying}.

\paragraph{Precision of H1.} For each H1 pair of the three leading zero-shot systems, the stored resamples give the distribution of the unweighted mean difference over test sets. With test sets $m$ times larger, drawn from the same population, the deviations of this mean from its centre shrink by $1/\sqrt{m}$. For a true difference $d$, we draw the estimate as $d$ plus a bootstrap deviation scaled by $1/\sqrt{m}$, build the 90\% percentile CI with the scaled quantiles, and count the draws in which the CI lies within $\pm0.02$. The smallest $m$ with a share of at least 0.80 is 4.0--4.4 against tabularisai\textsubscript{in} and 6.0--6.6 against XLM-EMO\textsubscript{in} for $d=0$, and 9.0--11.2 and 13.4--22.6 at the observed differences. The calculation assumes that larger test sets would come from the same sources, and it is post hoc.

\paragraph{Permutation tests for H4.} The variant test checks whether the differences between variants are larger than expected if the variant had no effect, both over all test sets (Holm over the five systems) and for each test set (Holm over the 50 combinations of system and test set). The position test measures the share of answers given to the option shown in each of the four positions; under the Latin square, if the position does not matter, each position receives 25\% of the answers. For each text, the four predictions of a system (one per variant) are permuted at random across the variants; the statistic is recomputed on each of $P=5{,}000$ permutations, and $p=(c+1)/(P+1)$, where $c$ counts the permutations with a statistic at least as large as the observed one. For the variant effect the statistic is the mean SD of macro-F1 across variants; for position bias it is the sum, over the four positions, of the squared deviations of the share of answers from 25\%.

\paragraph{Prompt and run components (complementary).} For Jev and GPT-6 Luna, each variant $v$ was run twice, with macro-F1 $F_{1v}$ and $F_{2v}$ on a test set. The variance due to the run is estimated as $s^2_{\text{run}} = \frac{1}{4}\sum_v (F_{1v}-F_{2v})^2/2$, and the variance due to the prompt as $s^2_{\text{prompt}} = \max\bigl(0,\, s^2_{\text{between}} - s^2_{\text{run}}/2\bigr)$, where $s^2_{\text{between}}$ is the variance across the four variants of the mean of the two runs. The first factor $1/2$ arises because the difference between two independent runs varies twice as much as a single run; the second because the mean of two runs keeps half of the run-to-run variance, which is removed so that only the variation due to the prompt remains. Both quantities are computed on each test set; their square roots, the two SDs, are averaged over test sets, and their CIs come from the same bootstrap resamples. The estimate assumes that the run-to-run variation is the same for the four variants; the two runs are about ten days apart, so any change in the service between them is counted as run-to-run variation; and with only two runs per variant both SDs are estimated imprecisely.

\paragraph{Calibration (complementary).} For a text $i$, let $p_{ic}$ be the probability assigned to class $c$, $\hat{y}_i$ the predicted class and $\text{conf}_i = p_{i\hat{y}_i}$ its probability. The expected calibration error groups the texts into 15 equal-width bins of $\text{conf}_i$ and is $\text{ECE} = \sum_b (n_b/n)\,|\text{acc}_b - \text{conf}_b|$, where $n_b$ is the number of texts in bin $b$ and $\text{acc}_b$ and $\text{conf}_b$ are their accuracy and mean confidence. The multiclass Brier score is $\frac{1}{n}\sum_i \sum_c (p_{ic} - y_{ic})^2$, with $y_{ic}=1$ for the gold class and 0 otherwise; it ranges from 0 to 2. Both are computed for each variant or model and averaged as macro-F1 is.

\emph{Class weights.} The encoders were trained with weights $w_c = N/(4\,n_c)$ (Section~\ref{sec:encoders}); their effect is removed by dividing each probability by $w_c$ and renormalising over the four classes, which is equivalent to subtracting $\log w_c$ from the logits. The largest weight is 147, for \emph{fear} in Universal Joy German.

\emph{Temperature scaling.} The logits are recovered as $z_{ic} = \log p_{ic}$, up to a constant per text that the softmax ignores; Jev and OpenAI Decisions round their probabilities to two decimals, so their probabilities of zero are replaced by 0.005, half the rounding unit, while the other systems never return a probability of zero. The recalibrated probabilities are $\text{softmax}(z_i/T)$, where the calibration temperature $T$ is estimated separately for each model or variant and test set by minimising the log-loss. Because predictions on the validation splits were not stored, each test set is split at random into two halves, with groups of identical texts kept together: $T$ is estimated on one half and applied to the other, and vice versa, and ECE and Brier are computed on the held-out halves. Temperature scaling does not change the predicted class.

\emph{Sensitivity.} Over 200 random splits into halves, the 2.5--97.5\% range of the mean expected calibration error is 0.043--0.053 for Jev, 0.037--0.046 for OpenAI Decisions, 0.035--0.042 for Strands Decider 2B, 0.056--0.060 and 0.040--0.046 for label-scored Qwen3.5-2B and Qwen3.5-4B, 0.046--0.058 for the encoders in-domain and 0.057--0.065 out of domain, so that Strands Decider 2B remains better calibrated than label-scored Qwen3.5-2B. With probabilities of zero replaced by 0.001, 0.0025 or 0.01 instead of 0.005, the recalibrated expected calibration error is 0.038--0.051 for Jev and 0.037--0.047 for OpenAI Decisions. For the encoders out of domain, a calibration temperature estimated on the in-domain test set of the same model, which needs no labels from the target domain (mean 2.1), gives 0.120 for tabularisai and 0.101 for XLM-EMO.

\emph{Bootstrap of the differences.} The comparisons (Strands Decider 2B against the two label-scored Qwen models; each decision model against each encoder, in-domain and out of domain), the measures and the decision rule were written down on 4 November 2026, after the point estimates were known and before any interval was computed. The resamples are those of the paired group bootstrap of macro-F1 ($B=10{,}000$); in every resample, ECE and Brier are recomputed for every model or variant, averaged as above, and, for the recalibrated probabilities, the calibration temperature is re-estimated on the resampled texts of each half, on a grid of $\log T$ from $-3$ to 3 in steps of 0.025. Brier differences are tested with the bootstrap $p$-value and Holm's correction over the 28 tests (14 comparisons, as returned and recalibrated); the ECE, which depends on the binning, receives 95\% percentile intervals only. The plan also foresaw intervals for the ECE of each system, which are not reported: the ECE of a resample is biased upwards, so that for most systems the bootstrap distribution lies above the estimate; in the differences between two systems this bias largely cancels. Results are in Table~\ref{tab:calibration-tests}.

%% file: tables/h1_per_dataset.tex
\begin{table*}[t]
\centering
\small
\setlength{\tabcolsep}{3.5pt}
\begin{tabular}{lccc}
\toprule
Test set & Jev $-$ tabularisai\textsubscript{in} & Jev $-$ XLM-EMO\textsubscript{in} & Train rows \\
\midrule
BRIGHTER de & +0.133$^{*}$ [0.10, 0.16] & +0.149$^{*}$ [0.12, 0.18] & 1,388 \\
EMit it & $-$0.044 [$-$0.12, 0.02] & $-$0.026 [$-$0.09, 0.04] & 1,432 \\
EmoEvent en & +0.020 [$-$0.03, 0.07] & +0.004 [$-$0.05, 0.06] & 1,995 \\
EmoEvent es & $-$0.004 [$-$0.04, 0.03] & 0.000 [$-$0.04, 0.04] & 2,569 \\
GoEmotions en & $-$0.181$^{*}$ [$-$0.21, $-$0.16] & $-$0.173$^{*}$ [$-$0.20, $-$0.15] & 24,802 \\
MultiEmotions-It & +0.064 [$-$0.16, 0.24] & +0.071 [$-$0.15, 0.24] & 1,305 \\
SemEval-18 en & +0.001 [$-$0.02, 0.02] & -- & 4,257 \\
SemEval-18 es & +0.032 [0.01, 0.05] & -- & 2,368 \\
Universal Joy de & $-$0.026 [$-$0.22, 0.02] & $-$0.067 [$-$0.28, $-$0.02] & 3,532 \\
Universal Joy fr & $-$0.082$^{*}$ [$-$0.21, $-$0.04] & $-$0.048 [$-$0.10, $-$0.01] & 3,800 \\
\bottomrule
\end{tabular}
\caption{H1 per test set: macro-F1 difference with group-bootstrap 95\% CI. $^{*}$: significant after Holm correction within the family; the CIs are not adjusted, so an interval that excludes zero can carry no asterisk. Train rows: size of the in-domain training split used by the fine-tuned encoder.}
\label{tab:h1-per-dataset}
\end{table*}

%% file: tables/h1_moderator.tex
\begin{table}[t]
\centering
\small
\setlength{\tabcolsep}{3pt}
\begin{adjustbox}{max width=\linewidth}
\begin{tabular}{lcccccccc}
\toprule
 & \multicolumn{4}{c}{All test sets} & \multicolumn{4}{c}{Without GoEmotions} \\
\cmidrule(lr){2-5}\cmidrule(lr){6-9}
 & \multicolumn{2}{c}{tabularisai\textsubscript{in}} & \multicolumn{2}{c}{XLM-EMO\textsubscript{in}} & \multicolumn{2}{c}{tabularisai\textsubscript{in}} & \multicolumn{2}{c}{XLM-EMO\textsubscript{in}} \\
\cmidrule(lr){2-3}\cmidrule(lr){4-5}\cmidrule(lr){6-7}\cmidrule(lr){8-9}
Zero-shot system & $\rho$ & $p$ & $\rho$ & $p$ & $\rho$ & $p$ & $\rho$ & $p$ \\
\midrule
Jev & $-$0.71 & 0.022 & $-$0.88 & 0.004 & $-$0.60 & 0.088 & $-$0.82 & 0.023 \\
OpenAI Decisions & $-$0.62 & 0.054 & $-$0.86 & 0.007 & $-$0.48 & 0.187 & $-$0.79 & 0.036 \\
GPT-6 Luna & $-$0.81 & 0.005 & $-$0.86 & 0.007 & $-$0.73 & 0.025 & $-$0.79 & 0.036 \\
Strands 2B & $-$0.22 & 0.533 & $-$0.45 & 0.260 & 0.07 & 0.865 & $-$0.18 & 0.702 \\
Qwen3.5-2B & $-$0.08 & 0.829 & $-$0.31 & 0.456 & 0.13 & 0.732 & $-$0.25 & 0.589 \\
\bottomrule
\end{tabular}
\end{adjustbox}
\caption{Exploratory (post hoc) Spearman correlation $\rho$ between the size of the in-domain training split and the per-dataset difference zero-shot minus fine-tuned, over 10 test sets against tabularisai\textsubscript{in} and 8 against XLM-EMO\textsubscript{in}. A negative $\rho$ means that the advantage of fine-tuning grows with the size of the training split. $p$: two-sided, from the $t$ approximation, with no multiplicity correction. GoEmotions has by far the largest training split (24{,}802 texts); the right-hand columns repeat the correlation without it (9 and 7 test sets).}
\label{tab:moderator}
\end{table}

%% file: tables/h1_aggregation.tex
\begin{table*}[t]
\centering
\small
\setlength{\tabcolsep}{4pt}
\begin{adjustbox}{max width=\linewidth}
\begin{tabular}{llccccc}
\toprule
Zero-shot & Reference & Unweighted (primary) & Row-weighted & Fixed effect & Pooled rows & Random effects \\
\midrule
Jev & tabularisai\textsubscript{in} & $-$0.009 [$-$0.045, 0.010] & $-$0.013 [$-$0.027, $-$0.003] & $-$0.009 [$-$0.020, 0.001] & $-$0.004 [$-$0.015, 0.008] & $-$0.011 [$-$0.074, 0.052] \\
OpenAI Decisions & tabularisai\textsubscript{in} & $-$0.010 [$-$0.044, 0.009] & $-$0.021 [$-$0.036, $-$0.012] & $-$0.022 [$-$0.032, $-$0.011] & $-$0.013 [$-$0.025, $-$0.002] & $-$0.010 [$-$0.074, 0.053] \\
GPT-6 Luna & tabularisai\textsubscript{in} & $-$0.010 [$-$0.045, 0.008] & $-$0.022 [$-$0.037, $-$0.012] & $-$0.022 [$-$0.032, $-$0.012] & $-$0.014 [$-$0.026, $-$0.003] & $-$0.011 [$-$0.078, 0.056] \\
Strands 2B & tabularisai\textsubscript{in} & $-$0.048 [$-$0.079, $-$0.033] & $-$0.031 [$-$0.046, $-$0.021] & $-$0.024 [$-$0.034, $-$0.014] & $-$0.024 [$-$0.035, $-$0.012] & $-$0.041 [$-$0.085, 0.003] \\
Qwen3.5-2B & tabularisai\textsubscript{in} & $-$0.145 [$-$0.166, $-$0.129] & $-$0.157 [$-$0.169, $-$0.146] & $-$0.134 [$-$0.145, $-$0.125] & $-$0.155 [$-$0.167, $-$0.143] & $-$0.147 [$-$0.184, $-$0.111] \\
\midrule
Jev & XLM-EMO\textsubscript{in} & $-$0.011 [$-$0.055, 0.012] & $-$0.018 [$-$0.039, $-$0.004] & $-$0.032 [$-$0.047, $-$0.017] & 0.000 [$-$0.018, 0.017] & $-$0.014 [$-$0.097, 0.068] \\
OpenAI Decisions & XLM-EMO\textsubscript{in} & $-$0.009 [$-$0.051, 0.014] & $-$0.025 [$-$0.045, $-$0.011] & $-$0.038 [$-$0.054, $-$0.024] & $-$0.009 [$-$0.026, 0.008] & $-$0.009 [$-$0.093, 0.075] \\
GPT-6 Luna & XLM-EMO\textsubscript{in} & $-$0.010 [$-$0.051, 0.013] & $-$0.026 [$-$0.046, $-$0.013] & $-$0.037 [$-$0.052, $-$0.024] & $-$0.011 [$-$0.028, 0.005] & $-$0.011 [$-$0.100, 0.079] \\
Strands 2B & XLM-EMO\textsubscript{in} & $-$0.058 [$-$0.095, $-$0.038] & $-$0.040 [$-$0.060, $-$0.027] & $-$0.044 [$-$0.060, $-$0.031] & $-$0.026 [$-$0.043, $-$0.009] & $-$0.050 [$-$0.103, 0.002] \\
Qwen3.5-2B & XLM-EMO\textsubscript{in} & $-$0.157 [$-$0.183, $-$0.135] & $-$0.185 [$-$0.200, $-$0.170] & $-$0.181 [$-$0.196, $-$0.166] & $-$0.194 [$-$0.211, $-$0.177] & $-$0.169 [$-$0.199, $-$0.139] \\
\bottomrule
\end{tabular}
\end{adjustbox}
\caption{H1 under different ways of aggregating the test sets: macro-F1 difference with 95\% CI. Unweighted (primary): mean of the per-test-set differences, each test set with the same weight. Row-weighted: the same mean, weighted by the number of test texts. Fixed effect: mean weighted by the inverse of the bootstrap variance of each test set. Pooled rows: macro-F1 computed once on all test texts pooled into a single set. Random effects: as in Table~\ref{tab:h1} (REML, Hartung--Knapp CI). CIs are from the group bootstrap except in the random-effects column. Row-weighted and pooled answer a different question, performance on the typical text of the collection, dominated by GoEmotions and BRIGHTER; the fixed-effect CI is too narrow given the heterogeneity ($I^2\approx0.97$).}
\label{tab:aggregation}
\end{table*}

%% file: tables/h1_no_uj.tex
\begin{table*}[t]
\centering
\small
\setlength{\tabcolsep}{4pt}
\begin{adjustbox}{max width=\linewidth}
\begin{tabular}{lllccccccllcccc}
\toprule
 & & & & \multicolumn{5}{c}{Mean over test sets without UJ} & & & \multicolumn{4}{c}{Random effects} \\
\cmidrule(lr){6-10}\cmidrule(lr){12-15}
Hyp. & A & B & $n$ & $\Delta$ all & $\Delta$ & 90\% CI & $\delta_{\min}$ & $p_{\text{Holm}}$ & TOST 0.02 & TOST 0.01 & $p_{\text{Holm}}$ & $\tau$ & $I^2$ & 95\% PI \\
\midrule
H1 & Jev & tabularisai\textsubscript{in} & 8 & $-$0.009 & +0.003 & [$-$0.023, 0.024] & 0.024 & 1.000 & inc. & inc. & 1.000 & 0.091 & 0.98 & [$-$0.24, 0.24] \\
 & OpenAI Decisions & tabularisai\textsubscript{in} & 8 & $-$0.010 & $-$0.005 & [$-$0.028, 0.016] & 0.028 & 1.000 & inc. & inc. & 1.000 & 0.096 & 0.98 & [$-$0.26, 0.24] \\
 & GPT-6 Luna & tabularisai\textsubscript{in} & 8 & $-$0.010 & $-$0.005 & [$-$0.028, 0.016] & 0.028 & 1.000 & inc. & inc. & 1.000 & 0.102 & 0.98 & [$-$0.27, 0.26] \\
 & Strands 2B & tabularisai\textsubscript{in} & 8 & $-$0.048 & $-$0.039 & [$-$0.057, $-$0.022] & 0.057 & 0.003 & diff. & diff. & 1.000 & 0.060 & 0.94 & [$-$0.19, 0.12] \\
 & Qwen3.5-2B & tabularisai\textsubscript{in} & 8 & $-$0.145 & $-$0.142 & [$-$0.160, $-$0.125] & 0.160 & 0.002 & diff. & diff. & 0.001 & 0.054 & 0.94 & [$-$0.29, $-$0.003] \\
 & Jev & XLM-EMO\textsubscript{in} & 6 & $-$0.011 & +0.004 & [$-$0.029, 0.032] & 0.032 & 1.000 & inc. & inc. & 1.000 & 0.109 & 0.98 & [$-$0.33, 0.33] \\
 & OpenAI Decisions & XLM-EMO\textsubscript{in} & 6 & $-$0.009 & $-$0.002 & [$-$0.031, 0.025] & 0.031 & 1.000 & inc. & inc. & 1.000 & 0.114 & 0.98 & [$-$0.35, 0.34] \\
 & GPT-6 Luna & XLM-EMO\textsubscript{in} & 6 & $-$0.010 & $-$0.002 & [$-$0.032, 0.025] & 0.032 & 1.000 & inc. & inc. & 1.000 & 0.122 & 0.99 & [$-$0.37, 0.36] \\
 & Strands 2B & XLM-EMO\textsubscript{in} & 6 & $-$0.058 & $-$0.048 & [$-$0.071, $-$0.025] & 0.071 & 0.004 & diff. & diff. & 1.000 & 0.066 & 0.94 & [$-$0.24, 0.15] \\
 & Qwen3.5-2B & XLM-EMO\textsubscript{in} & 6 & $-$0.157 & $-$0.156 & [$-$0.179, $-$0.134] & 0.179 & 0.002 & diff. & diff. & 0.001 & 0.028 & 0.60 & [$-$0.26, $-$0.08] \\
\midrule
H2 & tabularisai\textsubscript{in} & tabularisai\textsubscript{out} & 8 & +0.163 & +0.170 & [0.155, 0.187] & 0.187 & 0.002 & diff. & diff. & $<$0.001 & 0.043 & 0.93 & [0.07, 0.29] \\
 & XLM-EMO\textsubscript{in} & XLM-EMO\textsubscript{out} & 6 & +0.095 & +0.090 & [0.070, 0.111] & 0.111 & 0.002 & diff. & diff. & 0.016 & 0.041 & 0.89 & [$-$0.03, 0.22] \\
 & Jev & tabularisai\textsubscript{out} & 8 & +0.155 & +0.173 & [0.156, 0.188] & 0.188 & 0.002 & diff. & diff. & 0.001 & 0.064 & 0.97 & [0.01, 0.34] \\
 & OpenAI Decisions & tabularisai\textsubscript{out} & 8 & +0.154 & +0.165 & [0.151, 0.180] & 0.180 & 0.002 & diff. & diff. & 0.002 & 0.069 & 0.97 & [$-$0.01, 0.35] \\
 & GPT-6 Luna & tabularisai\textsubscript{out} & 8 & +0.153 & +0.166 & [0.151, 0.180] & 0.180 & 0.002 & diff. & diff. & 0.004 & 0.077 & 0.98 & [$-$0.03, 0.37] \\
 & Strands 2B & tabularisai\textsubscript{out} & 8 & +0.115 & +0.131 & [0.120, 0.142] & 0.142 & 0.002 & diff. & diff. & 0.004 & 0.059 & 0.92 & [$-$0.02, 0.29] \\
 & Jev & XLM-EMO\textsubscript{out} & 6 & +0.083 & +0.094 & [0.072, 0.113] & 0.113 & 0.002 & diff. & diff. & 0.261 & 0.095 & 0.98 & [$-$0.19, 0.38] \\
 & OpenAI Decisions & XLM-EMO\textsubscript{out} & 6 & +0.086 & +0.088 & [0.068, 0.108] & 0.108 & 0.002 & diff. & diff. & 0.261 & 0.095 & 0.98 & [$-$0.20, 0.38] \\
 & GPT-6 Luna & XLM-EMO\textsubscript{out} & 6 & +0.085 & +0.088 & [0.068, 0.108] & 0.108 & 0.002 & diff. & diff. & 0.261 & 0.104 & 0.99 & [$-$0.22, 0.40] \\
 & Strands 2B & XLM-EMO\textsubscript{out} & 6 & +0.037 & +0.042 & [0.029, 0.056] & 0.056 & 0.002 & diff. & diff. & 0.261 & 0.067 & 0.94 & [$-$0.16, 0.25] \\
\midrule
H3 & OpenAI Decisions & GPT-6 Luna & 8 & +0.001 & 0.000 & [$-$0.007, 0.007] & 0.007 & 1.000 & eq. & eq. & 0.788 & 0.010 & 0.64 & [$-$0.03, 0.03] \\
 & Jev & GPT-6 Luna & 8 & +0.001 & +0.007 & [$-$0.003, 0.017] & 0.017 & 0.476 & eq. & inc. & 0.555 & 0.017 & 0.74 & [$-$0.04, 0.06] \\
 & Strands 2B & Qwen3.5-2B & 8 & +0.097 & +0.103 & [0.090, 0.115] & 0.115 & $<$0.001 & diff. & diff. & 0.028 & 0.081 & 0.97 & [$-$0.11, 0.31] \\
\bottomrule
\end{tabular}
\end{adjustbox}
\caption{H1, H2 and H3 without the two Universal Joy test sets (UJ), in which the class \emph{fear} has 1 and 2 test texts. $\Delta$ all: primary analysis; the other columns are recomputed on the remaining test sets with the same resamples. $\delta_{\min}$: smallest equivalence margin, the larger absolute bound of the 90\% CI. TOST outcomes: eq., equivalent; diff., different; inc., inconclusive. $p_{\text{Holm}}$ within each family; random effects as in Table~\ref{tab:h1}. A bound that would round to zero is shown with one more decimal, to keep its side of zero.}
\label{tab:no-uj}
\end{table*}

%% file: tables/wilcoxon.tex
\begin{table*}[t]
\centering
\small
\begin{adjustbox}{max width=\linewidth}
\begin{tabular}{lllcccc}
\toprule
Hypothesis & A & B & $n$ & Wilcoxon $p$ & $p_{\text{Holm}}$ & smallest attainable $p$ \\
\midrule
H1 & Jev & tabularisai\textsubscript{in} & 10 & 0.846 & 1.000 & 0.00195 \\
 & OpenAI Decisions & tabularisai\textsubscript{in} & 10 & 0.846 & 1.000 & 0.00195 \\
 & GPT-6 Luna & tabularisai\textsubscript{in} & 10 & 1.000 & 1.000 & 0.00195 \\
 & Strands 2B & tabularisai\textsubscript{in} & 10 & 0.049 & 0.342 & 0.00195 \\
 & Qwen3.5-2B & tabularisai\textsubscript{in} & 10 & 0.002 & 0.020 & 0.00195 \\
 & Jev & XLM-EMO\textsubscript{in} & 8 & 0.844 & 1.000 & 0.00781 \\
 & OpenAI Decisions & XLM-EMO\textsubscript{in} & 8 & 0.844 & 1.000 & 0.00781 \\
 & GPT-6 Luna & XLM-EMO\textsubscript{in} & 8 & 0.945 & 1.000 & 0.00781 \\
 & Strands 2B & XLM-EMO\textsubscript{in} & 8 & 0.039 & 0.312 & 0.00781 \\
 & Qwen3.5-2B & XLM-EMO\textsubscript{in} & 8 & 0.008 & 0.070 & 0.00781 \\
\midrule
H2 & tabularisai\textsubscript{in} & tabularisai\textsubscript{out} & 10 & 0.002 & 0.020 & 0.00195 \\
 & XLM-EMO\textsubscript{in} & XLM-EMO\textsubscript{out} & 8 & 0.008 & 0.039 & 0.00781 \\
 & Jev & tabularisai\textsubscript{out} & 10 & 0.002 & 0.020 & 0.00195 \\
 & OpenAI Decisions & tabularisai\textsubscript{out} & 10 & 0.002 & 0.020 & 0.00195 \\
 & GPT-6 Luna & tabularisai\textsubscript{out} & 10 & 0.002 & 0.020 & 0.00195 \\
 & Strands 2B & tabularisai\textsubscript{out} & 10 & 0.002 & 0.020 & 0.00195 \\
 & Jev & XLM-EMO\textsubscript{out} & 8 & 0.016 & 0.062 & 0.00781 \\
 & OpenAI Decisions & XLM-EMO\textsubscript{out} & 8 & 0.016 & 0.062 & 0.00781 \\
 & GPT-6 Luna & XLM-EMO\textsubscript{out} & 8 & 0.023 & 0.062 & 0.00781 \\
 & Strands 2B & XLM-EMO\textsubscript{out} & 8 & 0.055 & 0.062 & 0.00781 \\
\midrule
H3 & OpenAI Decisions & GPT-6 Luna & 10 & 0.922 & 1.000 & 0.00195 \\
 & Jev & GPT-6 Luna & 10 & 0.846 & 1.000 & 0.00195 \\
 & Strands 2B & Qwen3.5-2B & 10 & 0.002 & 0.006 & 0.00195 \\
\bottomrule
\end{tabular}
\end{adjustbox}
\caption{Question 3 of every comparison: exact Wilcoxon signed-rank test on the per-dataset differences, two-sided, Holm-adjusted within each family. The smallest attainable $p$ (all differences with the same sign) is $2/2^{10}=0.00195$ with 10 test sets and $2/2^{8}=0.00781$ with 8, so with 8 test sets the test can survive Holm correction only if few pairs of its family have a smaller $p$, as for XLM-EMO\textsubscript{in} $-$ XLM-EMO\textsubscript{out}.}
\label{tab:wilcoxon}
\end{table*}

%% file: tables/h4_extra.tex
\begin{table*}[t]
\centering
\small
\setlength{\tabcolsep}{4pt}
\begin{adjustbox}{max width=\linewidth}
\begin{tabular}{llcccccc}
\toprule
A & B & $n$ & $\Delta$ & 95\% CI & $p_{\text{Holm}}$ & $\Delta_{\text{RE}}$ [95\% CI] & RE $p_{\text{Holm}}$ \\
\midrule
tabularisai\textsubscript{out, same lang.} & tabularisai\textsubscript{out, other lang.} & 9 & +0.042 & [0.031, 0.052] & 0.004 & +0.042 [$-$0.017, 0.100] & 0.643 \\
XLM-EMO\textsubscript{out, same lang.} & XLM-EMO\textsubscript{out, other lang.} & 6 & $-$0.004 & [$-$0.020, 0.012] & 0.627 & $-$0.007 [$-$0.046, 0.031] & 0.646 \\
Jev & tabularisai\textsubscript{out, same lang.} & 9 & +0.128 & [0.109, 0.148] & 0.004 & +0.130 [0.042, 0.217] & 0.127 \\
Jev & tabularisai\textsubscript{out, other lang.} & 9 & +0.170 & [0.151, 0.186] & 0.004 & +0.171 [0.122, 0.220] & $<$0.001 \\
Jev & XLM-EMO\textsubscript{out, same lang.} & 6 & +0.097 & [0.069, 0.124] & 0.004 & +0.097 [$-$0.023, 0.218] & 0.643 \\
Jev & XLM-EMO\textsubscript{out, other lang.} & 6 & +0.093 & [0.067, 0.116] & 0.004 & +0.091 [$-$0.007, 0.190] & 0.569 \\
\midrule
Jev & tabularisai\textsubscript{out, best} & 10 & +0.071 & [0.046, 0.088] & 0.002 & +0.073 [0.026, 0.121] & 0.056 \\
OpenAI Decisions & tabularisai\textsubscript{out, best} & 10 & +0.070 & [0.045, 0.087] & 0.002 & +0.072 [0.023, 0.121] & 0.063 \\
GPT-6 Luna & tabularisai\textsubscript{out, best} & 10 & +0.070 & [0.045, 0.086] & 0.002 & +0.071 [0.016, 0.125] & 0.101 \\
Strands 2B & tabularisai\textsubscript{out, best} & 10 & +0.032 & [0.009, 0.046] & 0.010 & +0.038 [$-$0.004, 0.081] & 0.355 \\
Jev & XLM-EMO\textsubscript{out, best} & 8 & +0.048 & [0.026, 0.068] & 0.002 & +0.046 [$-$0.021, 0.113] & 0.459 \\
OpenAI Decisions & XLM-EMO\textsubscript{out, best} & 8 & +0.050 & [0.029, 0.070] & 0.002 & +0.050 [$-$0.016, 0.116] & 0.459 \\
GPT-6 Luna & XLM-EMO\textsubscript{out, best} & 8 & +0.049 & [0.029, 0.069] & 0.002 & +0.049 [$-$0.024, 0.121] & 0.459 \\
Strands 2B & XLM-EMO\textsubscript{out, best} & 8 & +0.002 & [$-$0.015, 0.017] & 0.871 & +0.005 [$-$0.044, 0.053] & 0.827 \\
tabularisai\textsubscript{in} & tabularisai\textsubscript{out, best} & 10 & +0.080 & [0.061, 0.108] & 0.002 & +0.088 [0.056, 0.119] & 0.001 \\
XLM-EMO\textsubscript{in} & XLM-EMO\textsubscript{out, best} & 8 & +0.059 & [0.039, 0.096] & 0.002 & +0.060 [0.023, 0.097] & 0.056 \\
\bottomrule
\end{tabular}
\end{adjustbox}
\caption{H2, complementary analyses. Top: out-of-domain sources split by language (same language as the test set vs.\ other languages; only test sets with both kinds of source). Bottom: the best out-of-domain source for each test set, that is, the source dataset with the highest mean over its three seeds on the test set itself (an oracle choice that favours the fine-tuned encoder; the choice is kept fixed in the bootstrap). Holm within each block (18 and 10 pairs, including pairs for OpenAI Decisions, GPT-6 Luna and Strands 2B in the top block, omitted here and available in the released files).}
\label{tab:h4-extra}
\end{table*}

%% file: tables/exploratory.tex
\begin{table}[t]
\centering
\small
\setlength{\tabcolsep}{3pt}
\begin{adjustbox}{max width=\linewidth}
\begin{tabular}{llccccclcc}
\toprule
A & B & $n$ & $\Delta$ & 95\% CI & 90\% CI & $p_{\text{Holm}}$ & TOST ($\delta{=}0.02$) & RE $p_{\text{Holm}}$ & RE 95\% PI \\
\midrule
tabularisai\textsubscript{in} & XLM-EMO\textsubscript{in} & 8 & +0.004 & [$-$0.011, 0.022] & [$-$0.008, 0.018] & 0.575 & equivalent & 0.202 & [$-$0.005, 0.018] \\
tabularisai\textsubscript{out} & XLM-EMO\textsubscript{out} & 8 & $-$0.059 & [$-$0.069, $-$0.047] & [$-$0.067, $-$0.049] & $<$0.001 & different & 0.005 & [$-$0.144, 0.027] \\
\midrule
Jev & tabularisai\textsubscript{zs} & 10 & +0.183 & [0.163, 0.200] & [0.166, 0.197] & 0.002 & different & $<$0.001 & [0.083, 0.294] \\
OpenAI Decisions & tabularisai\textsubscript{zs} & 10 & +0.182 & [0.164, 0.199] & [0.167, 0.196] & 0.002 & different & $<$0.001 & [0.098, 0.281] \\
GPT-6 Luna & tabularisai\textsubscript{zs} & 10 & +0.181 & [0.164, 0.197] & [0.166, 0.195] & 0.002 & different & $<$0.001 & [0.085, 0.288] \\
Strands 2B & tabularisai\textsubscript{zs} & 10 & +0.143 & [0.127, 0.159] & [0.129, 0.156] & 0.002 & different & $<$0.001 & [$-$0.012, 0.311] \\
Jev & XLM-EMO\textsubscript{zs} & 8 & +0.063 & [0.041, 0.086] & [0.044, 0.082] & 0.002 & different & 0.285 & [$-$0.150, 0.274] \\
OpenAI Decisions & XLM-EMO\textsubscript{zs} & 8 & +0.065 & [0.044, 0.088] & [0.048, 0.084] & 0.002 & different & 0.285 & [$-$0.157, 0.288] \\
GPT-6 Luna & XLM-EMO\textsubscript{zs} & 8 & +0.064 & [0.042, 0.087] & [0.046, 0.084] & 0.002 & different & 0.285 & [$-$0.176, 0.305] \\
Strands 2B & XLM-EMO\textsubscript{zs} & 8 & +0.017 & [$-$0.001, 0.035] & [0.002, 0.032] & 0.068 & inconclusive & 0.370 & [$-$0.121, 0.160] \\
\bottomrule
\end{tabular}
\end{adjustbox}
\caption{Exploratory comparisons. Top: the two fine-tuned encoders against each other (8 test sets). Bottom: zero-shot systems against the public encoders used without any training on our data (\textsubscript{zs}). Columns as in Tables~\ref{tab:h1} and~\ref{tab:h4}; $p$ values Holm-adjusted within each of the two groups. For tabularisai\textsubscript{in} $-$ XLM-EMO\textsubscript{in} the estimated between-dataset variance is zero, so the prediction interval is narrower than the bootstrap CI of the unweighted mean: the random-effects model weights the datasets by their precision.}
\label{tab:exploratory}
\end{table}

%% file: tables/per_class.tex
\begin{table*}[t]
\centering
\small
\setlength{\tabcolsep}{4pt}
\begin{adjustbox}{max width=\linewidth}
\begin{tabular}{lcccccccccccccc}
\toprule
 & \multicolumn{4}{c}{Mean over test sets} & \multicolumn{4}{c}{GoEmotions en} & \multicolumn{4}{c}{BRIGHTER de} & \multicolumn{2}{c}{Universal Joy, de / fr} \\
\cmidrule(lr){2-5}\cmidrule(lr){6-9}\cmidrule(lr){10-13}\cmidrule(lr){14-15}
System & joy & sad. & anger & fear & joy & sad. & anger & fear & joy & sad. & anger & fear & F1 \emph{fear} & macro-F1 w/o \emph{fear} \\
\midrule
Jev & 0.90 & 0.69 & 0.72 & 0.52 & 0.82 & 0.49 & 0.65 & 0.43 & 0.92 & 0.80 & 0.91 & 0.75 & 0.00 / 0.00 & 0.65 / 0.67 \\
OpenAI Decisions & 0.90 & 0.68 & 0.75 & 0.50 & 0.83 & 0.46 & 0.63 & 0.36 & 0.93 & 0.80 & 0.91 & 0.72 & 0.00 / 0.00 & 0.67 / 0.71 \\
GPT-6 Luna & 0.90 & 0.69 & 0.74 & 0.50 & 0.81 & 0.48 & 0.62 & 0.34 & 0.93 & 0.83 & 0.92 & 0.75 & 0.00 / 0.00 & 0.67 / 0.71 \\
Strands 2B & 0.87 & 0.64 & 0.70 & 0.46 & 0.87 & 0.49 & 0.69 & 0.55 & 0.85 & 0.75 & 0.84 & 0.59 & 0.00 / 0.00 & 0.59 / 0.65 \\
Qwen3.5-2B & 0.83 & 0.60 & 0.49 & 0.37 & 0.84 & 0.46 & 0.45 & 0.61 & 0.77 & 0.54 & 0.46 & 0.15 & 0.00 / 0.00 & 0.48 / 0.56 \\
\midrule
tabularisai\textsubscript{in} & 0.89 & 0.72 & 0.74 & 0.52 & 0.93 & 0.69 & 0.77 & 0.72 & 0.82 & 0.67 & 0.83 & 0.54 & 0.00 / 0.00 & 0.68 / 0.78 \\
XLM-EMO\textsubscript{in}$^{*}$ & 0.89 & 0.73 & 0.72 & 0.42 & 0.93 & 0.69 & 0.77 & 0.69 & 0.82 & 0.68 & 0.82 & 0.46 & 0.00 / 0.00 & 0.74 / 0.73 \\
\midrule
tabularisai\textsubscript{out} & 0.80 & 0.54 & 0.57 & 0.30 & 0.77 & 0.38 & 0.54 & 0.37 & 0.71 & 0.53 & 0.71 & 0.30 & 0.01 / 0.01 & 0.54 / 0.55 \\
XLM-EMO\textsubscript{out}$^{*}$ & 0.84 & 0.59 & 0.62 & 0.33 & 0.83 & 0.48 & 0.64 & 0.47 & 0.76 & 0.56 & 0.73 & 0.30 & 0.01 / 0.04 & 0.57 / 0.59 \\
\bottomrule
\end{tabular}
\end{adjustbox}
\caption{F1 of each class, averaged over prompt variants or seeds (and, for \textsubscript{out}, over source datasets). Mean over test sets: unweighted, over 10 test sets ($^{*}$8 for XLM-EMO). GoEmotions en: test support 2{,}034 \emph{joy}, 320 \emph{sadness}, 661 \emph{anger} and 89 \emph{fear} texts. BRIGHTER de: the dataset where the zero-shot systems are most ahead of in-domain fine-tuning. Universal Joy, F1 \emph{fear}: F1 of \emph{fear} on the German and French test sets, which contain 1 and 2 \emph{fear} test texts; none of the zero-shot systems and in-domain encoders recognises them (F1 of 0) and the out-of-domain encoders rarely do, so \emph{fear} contributes an F1 of zero or close to zero to the macro-F1 of Universal Joy. Macro-F1 w/o \emph{fear}: macro-F1 on Universal Joy over the other three classes only, 0.10--0.19 higher than over four classes (Table~\ref{tab:overview}).}
\label{tab:per-class}
\end{table*}

%% file: tables/complementary.tex
\begin{table*}[t]
\centering
\small
\setlength{\tabcolsep}{4pt}
\begin{adjustbox}{max width=\linewidth}
\begin{tabular}{lcccccc}
\toprule
Comparison & $n$ & $\Delta$ & 90\% CI & $p_{\text{Holm}}$ & TOST ($\delta{=}0.02$) & TOST ($\delta{=}0.01$) \\
\midrule
\multicolumn{7}{l}{\emph{(a) Label scoring (LS) of Qwen3.5-2B, Italian instructions}} \\
Strands 2B $-$ Qwen3.5-2B (as in Table~\ref{tab:h2}) & 10 & +0.097 & [0.078, 0.106] & 0.001 & different & different \\
Qwen LS $-$ Qwen generative (format) & 10 & +0.001 & [$-$0.0003, 0.003] & 0.190 & equivalent & equivalent \\
Strands 2B $-$ Qwen LS (adapter and distillation) & 10 & +0.096 & [0.077, 0.105] & 0.001 & different & different \\
Qwen LS-mean $-$ Qwen generative & 10 & $-$0.012 & [$-$0.021, $-$0.008] & 0.001 & different & different \\
Strands 2B $-$ Qwen LS-mean & 10 & +0.109 & [0.094, 0.118] & 0.001 & different & different \\
\midrule
\multicolumn{7}{l}{\emph{(b) Label scoring of Qwen3.5-4B, the model used as teacher for Strands Decider 2B, Italian instructions}} \\
Qwen3.5-4B LS $-$ Qwen3.5-2B LS (scale) & 10 & +0.093 & [0.074, 0.102] & $<$0.001 & different & different \\
Strands 2B $-$ Qwen3.5-4B LS & 10 & +0.003 & [$-$0.008, 0.013] & 1.000 & equivalent & inconclusive \\
Strands 2B $-$ Qwen3.5-4B LS-mean & 10 & +0.004 & [$-$0.007, 0.015] & 1.000 & equivalent & inconclusive \\
Jev $-$ Qwen3.5-4B LS & 10 & +0.042 & [0.031, 0.052] & $<$0.001 & different & different \\
\midrule
\multicolumn{7}{l}{\emph{(c) Generative Qwen3.5-4B, Italian instructions}} \\
Qwen3.5-4B LS $-$ Qwen3.5-4B generative (format) & 10 & +0.001 & [$-$0.001, 0.002] & 0.976 & equivalent & equivalent \\
Qwen3.5-4B $-$ Qwen3.5-2B, generative (scale) & 10 & +0.094 & [0.075, 0.102] & $<$0.001 & different & different \\
Strands 2B $-$ Qwen3.5-4B generative & 10 & +0.003 & [$-$0.008, 0.014] & 0.976 & equivalent & inconclusive \\
Jev $-$ Qwen3.5-4B generative & 10 & +0.043 & [0.031, 0.052] & $<$0.001 & different & different \\
\midrule
\multicolumn{7}{l}{\emph{(d) Instructions in English}} \\
Strands 2B: English $-$ Italian & 10 & $-$0.008 & [$-$0.018, $-$0.0003] & 0.270 & equivalent & inconclusive \\
Qwen generative: English $-$ Italian & 10 & $-$0.008 & [$-$0.019, $-$0.003] & 0.086 & equivalent & inconclusive \\
Qwen LS: English $-$ Italian & 10 & $-$0.004 & [$-$0.015, 0.001] & 0.384 & equivalent & inconclusive \\
Strands 2B $-$ Qwen generative, English & 10 & +0.097 & [0.080, 0.108] & 0.001 & different & different \\
Gap Strands $-$ Qwen: English $-$ Italian & 10 & 0.000 & [$-$0.009, 0.012] & 1.000 & equivalent & inconclusive \\
\midrule
\multicolumn{7}{l}{\emph{(e) Instructions in the language of the text (German, Spanish and French test sets): 2B models}} \\
Strands 2B $-$ Qwen generative, Italian & 5 & +0.130 & [0.098, 0.140] & 0.001 & different & different \\
Strands 2B: text language $-$ Italian & 5 & $-$0.004 & [$-$0.010, 0.002] & 0.282 & equivalent & equivalent \\
Qwen generative: text language $-$ Italian & 5 & +0.062 & [0.042, 0.071] & 0.001 & different & different \\
Strands 2B $-$ Qwen generative, text language & 5 & +0.064 & [0.040, 0.075] & 0.001 & different & different \\
Gap Strands $-$ Qwen: text language $-$ Italian & 5 & $-$0.066 & [$-$0.075, $-$0.046] & 0.001 & different & different \\
\midrule
\multicolumn{7}{l}{\emph{(f) Instructions in the language of the text (same test sets): leading systems and Qwen3.5-4B}} \\
Jev: text language $-$ Italian & 5 & +0.006 & [0.002, 0.011] & 0.167 & equivalent & inconclusive \\
OpenAI Decisions: text language $-$ Italian & 5 & +0.006 & [0.0003, 0.012] & 0.362 & equivalent & inconclusive \\
GPT-6 Luna: text language $-$ Italian & 5 & 0.000 & [$-$0.004, 0.004] & 1.000 & equivalent & equivalent \\
OpenAI Decisions $-$ GPT-6 Luna, text language & 5 & +0.005 & [$-$0.003, 0.012] & 0.943 & equivalent & inconclusive \\
Gap Decisions $-$ Luna: text language $-$ Italian & 5 & +0.006 & [$-$0.001, 0.013] & 0.741 & equivalent & inconclusive \\
Qwen3.5-4B generative: text language $-$ Italian & 5 & +0.015 & [0.008, 0.022] & 0.001 & different & different \\
Strands 2B $-$ Qwen3.5-4B generative, text language & 5 & $-$0.002 & [$-$0.013, 0.009] & 1.000 & equivalent & inconclusive \\
\midrule
\multicolumn{7}{l}{\emph{(g) Second run of the four variants}} \\
Jev: second run $-$ first run & 10 & 0.000 & [$-$0.002, 0.002] & 1.000 & equivalent & equivalent \\
GPT-6 Luna: second run $-$ first run & 10 & +0.003 & [$-$0.0002, 0.007] & 0.482 & equivalent & equivalent \\
OpenAI Decisions $-$ GPT-6 Luna (second run) & 10 & $-$0.003 & [$-$0.009, 0.004] & 1.000 & equivalent & equivalent \\
Jev $-$ GPT-6 Luna (both second runs) & 10 & $-$0.002 & [$-$0.011, 0.007] & 1.000 & equivalent & inconclusive \\
\bottomrule
\end{tabular}
\end{adjustbox}
\caption{Complementary runs: paired macro-F1 differences, unweighted mean over test sets, with the same group-bootstrap resamples as the main analysis; $p_{\text{Holm}}$ within each block. LS: label scoring, the summed log-probability of each label as continuation of the generative prompt; LS-mean: the same with the mean log-probability per token. Gap: difference of the two differences. In blocks (e) and (f) all rows use the same five test sets. CIs are not adjusted for multiplicity; $p_{\text{Holm}}$ is. A bound that would round to zero is shown with one more decimal.}
\label{tab:complementary}
\end{table*}

%% file: tables/complementary_h4.tex
\begin{table}[t]
\centering
\small
\setlength{\tabcolsep}{4pt}
\begin{adjustbox}{max width=\linewidth}
\begin{tabular}{lccccc}
\toprule
System and condition & $n$ & SD & UB\textsubscript{90} & max position (pp) & invalid \\
\midrule
\multicolumn{6}{l}{\emph{(a) All test sets}} \\
Qwen3.5-2B generative, Italian (as in Table~\ref{tab:h3}) & 10 & 0.073 & 0.0850 & +10.6 (pos.\,1) & 6 \\
Qwen3.5-2B LS, Italian & 10 & 0.074 & 0.0860 & +10.3 (pos.\,1) & 0 \\
Qwen3.5-4B LS, Italian & 10 & 0.024 & 0.0319 & +4.1 (pos.\,1) & 0 \\
Qwen3.5-4B generative, Italian & 10 & 0.025 & 0.0326 & +4.3 (pos.\,1) & 1 \\
Qwen3.5-2B generative, English & 10 & 0.073 & 0.0808 & +17.6 (pos.\,1) & 178 \\
Qwen3.5-2B LS, English & 10 & 0.068 & 0.0760 & +17.0 (pos.\,1) & 0 \\
Strands 2B, Italian (as in Table~\ref{tab:h3}) & 10 & 0.012 & 0.0204 & +1.4 (pos.\,1) & 0 \\
Strands 2B, English & 10 & 0.008 & 0.0139 & +0.8 (pos.\,4) & 0 \\
\midrule
\multicolumn{6}{l}{\emph{(b) German, Spanish and French test sets}} \\
Qwen3.5-2B generative, Italian & 5 & 0.075 & 0.0951 & +11.5 (pos.\,1) & 0 \\
Qwen3.5-2B generative, text language & 5 & 0.036 & 0.0481 & +3.9 (pos.\,1) & 0 \\
Strands 2B, Italian & 5 & 0.009 & 0.0146 & +1.8 (pos.\,1) & 0 \\
Strands 2B, text language & 5 & 0.012 & 0.0177 & +1.8 (pos.\,1) & 0 \\
Qwen3.5-4B generative, Italian & 5 & 0.017 & 0.0245 & +4.7 (pos.\,1) & 0 \\
Qwen3.5-4B generative, text language & 5 & 0.020 & 0.0264 & +4.8 (pos.\,1) & 0 \\
Jev, Italian & 5 & 0.007 & 0.0111 & $-$1.0 (pos.\,3) & 0 \\
Jev, text language & 5 & 0.006 & 0.0102 & $-$0.9 (pos.\,3) & 0 \\
OpenAI Decisions, Italian & 5 & 0.009 & 0.0129 & $-$0.5 (pos.\,4) & 0 \\
OpenAI Decisions, text language & 5 & 0.010 & 0.0171 & +1.0 (pos.\,1) & 2 \\
GPT-6 Luna, Italian & 5 & 0.009 & 0.0133 & $-$0.7 (pos.\,1) & 0 \\
GPT-6 Luna, text language & 5 & 0.010 & 0.0159 & +0.4 (pos.\,4) & 1 \\
\midrule
\multicolumn{6}{l}{\emph{(c) Second run of the four variants, all test sets}} \\
Jev, second run & 10 & 0.008 & 0.0134 & $-$1.0 (pos.\,3) & 0 \\
GPT-6 Luna, second run & 10 & 0.013 & 0.0198 & $-$0.7 (pos.\,1) & 7 \\
\midrule
\multicolumn{6}{l}{\emph{(d) Prompt and run components, from the two runs}} \\
 & \multicolumn{2}{c}{SD, prompt [95\% CI]} & \multicolumn{2}{c}{SD, run [95\% CI]} & same answer \\
Jev & \multicolumn{2}{c}{0.007 [0.006, 0.011]} & \multicolumn{2}{c}{0.004 [0.004, 0.008]} & 98.9\% \\
GPT-6 Luna & \multicolumn{2}{c}{0.008 [0.006, 0.012]} & \multicolumn{2}{c}{0.009 [0.009, 0.016]} & 95.4\% \\
\bottomrule
\end{tabular}
\end{adjustbox}
\caption{H4 measures for the complementary runs: mean SD of macro-F1 across the four variants with the upper bound of its 90\% CI (\emph{stable} if below 0.02), largest deviation of the share of answers given to a list position from 25\% (percentage points), and number of invalid answers (four answers per text, one per variant). Block (d): the variation across variants split into a part due to the prompt and a part due to the run (Section~\ref{sec:complementary}), and the share of answers that are the same in the two runs.}
\label{tab:complementary-h4}
\end{table}

%% file: tables/controlled_latency.tex
\begin{table}[t]
\centering
\small
\setlength{\tabcolsep}{4pt}
\begin{adjustbox}{max width=\linewidth}
\begin{tabular}{lccccccc}
\toprule
 & \multicolumn{4}{c}{Latency (ms)} & \multicolumn{2}{c}{Tokens per request} & USD per \\
\cmidrule(lr){2-5}\cmidrule(lr){6-7}
API & median & median, morning & median, afternoon & 95th pct. & input & output & 1{,}000 requests \\
\midrule
Jev & 156 & 152 & 159 & 178 & 442 & 48.2 & 0.0186$^{b}$ \\
OpenAI Decisions & 385 & 376 & 389 & 613 & 222 & 0.0 & 0.0222$^{c}$ \\
GPT-6 Luna & 2{,}069 & 2{,}014 & 2{,}124 & 3{,}530 & 164 & 5.2 & 0.0190$^{a}$ \\
\bottomrule
\end{tabular}
\end{adjustbox}
\caption{Controlled latency: 600 texts (60 per test set, variant 1), each sent to the three APIs one after the other in random order, with one request at a time, from the same client, in two time windows (3{,}600 requests, all valid). Ratio of the medians, with 95\% CI from a bootstrap over texts: GPT-6 Luna / OpenAI Decisions 5.38 [5.10, 5.64]; OpenAI Decisions / Jev 2.46 [2.37, 2.57]. Tokens as reported by each API; the tokenisers differ, so the token counts of Jev are not comparable with those of the OpenAI models. Cost from the token counts and list prices consulted on 30 October 2026: $^{a}$official OpenAI price list (USD 0.10 and 0.50 per million input and output tokens); $^{b}$input price from the provider's home page (USD 0.042 per million), output price of zero from third-party sources; $^{c}$not on the official OpenAI price list, input price of USD 0.10 per million and free output from third-party sources. Costs b and c are therefore unverified.}
\label{tab:controlled-latency}
\end{table}

%% file: tables/calibration_tests.tex
\begin{table*}[t]
\centering
\small
\setlength{\tabcolsep}{4pt}
\begin{adjustbox}{max width=\linewidth}
\begin{tabular}{llcccccc}
\toprule
 & & \multicolumn{3}{c}{As returned} & \multicolumn{3}{c}{Recalibrated} \\
\cmidrule(lr){3-5}\cmidrule(lr){6-8}
$A$ & $B$ & $\Delta$ECE [95\% CI] & $\Delta$Brier [95\% CI] & $p_{\text{Holm}}$ & $\Delta$ECE [95\% CI] & $\Delta$Brier [95\% CI] & $p_{\text{Holm}}$ \\
\midrule
Strands 2B & Qwen3.5-2B, LS & $-$0.016 [$-$0.023, $-$0.007] & $-$0.102 [$-$0.111, $-$0.093] & 0.006 & $-$0.021 [$-$0.027, $-$0.012] & $-$0.087 [$-$0.095, $-$0.079] & 0.006 \\
Strands 2B & Qwen3.5-4B, LS & +0.013 [0.005, 0.022] & $-$0.008 [$-$0.018, 0.001] & 0.598 & $-$0.008 [$-$0.014, 0.002] & $-$0.009 [$-$0.018, $-$0.0005] & 0.408 \\
\midrule
Jev & tabularisai\textsubscript{in} & $-$0.031 [$-$0.039, $-$0.018] & $-$0.019 [$-$0.038, 0.001] & 0.576 & $-$0.001 [$-$0.010, 0.015] & $-$0.002 [$-$0.017, 0.013] & 1.000 \\
Jev & tabularisai\textsubscript{out} & $-$0.128 [$-$0.134, $-$0.117] & $-$0.240 [$-$0.255, $-$0.226] & 0.006 & $-$0.013 [$-$0.023, $-$0.001] & $-$0.170 [$-$0.182, $-$0.159] & 0.006 \\
Jev & XLM-EMO\textsubscript{in} & $-$0.024 [$-$0.035, $-$0.009] & $-$0.003 [$-$0.027, 0.022] & 1.000 & $-$0.004 [$-$0.012, 0.019] & +0.019 [0.002, 0.037] & 0.392 \\
Jev & XLM-EMO\textsubscript{out} & $-$0.086 [$-$0.093, $-$0.074] & $-$0.137 [$-$0.155, $-$0.119] & 0.006 & $-$0.009 [$-$0.020, 0.003] & $-$0.086 [$-$0.100, $-$0.074] & 0.006 \\
OpenAI Decisions & tabularisai\textsubscript{in} & $-$0.040 [$-$0.047, $-$0.026] & $-$0.025 [$-$0.045, $-$0.005] & 0.151 & $-$0.006 [$-$0.014, 0.010] & $-$0.007 [$-$0.022, 0.008] & 1.000 \\
OpenAI Decisions & tabularisai\textsubscript{out} & $-$0.137 [$-$0.142, $-$0.126] & $-$0.247 [$-$0.262, $-$0.232] & 0.006 & $-$0.018 [$-$0.027, $-$0.006] & $-$0.175 [$-$0.187, $-$0.164] & 0.006 \\
OpenAI Decisions & XLM-EMO\textsubscript{in} & $-$0.034 [$-$0.044, $-$0.018] & $-$0.015 [$-$0.040, 0.010] & 1.000 & $-$0.009 [$-$0.015, 0.013] & +0.009 [$-$0.009, 0.027] & 1.000 \\
OpenAI Decisions & XLM-EMO\textsubscript{out} & $-$0.096 [$-$0.102, $-$0.083] & $-$0.149 [$-$0.167, $-$0.131] & 0.006 & $-$0.013 [$-$0.023, $-$0.003] & $-$0.097 [$-$0.111, $-$0.084] & 0.006 \\
Strands 2B & tabularisai\textsubscript{in} & $-$0.042 [$-$0.046, $-$0.025] & $-$0.005 [$-$0.021, 0.013] & 1.000 & $-$0.009 [$-$0.014, 0.005] & +0.021 [0.007, 0.034] & 0.042 \\
Strands 2B & tabularisai\textsubscript{out} & $-$0.139 [$-$0.142, $-$0.123] & $-$0.226 [$-$0.238, $-$0.215] & 0.006 & $-$0.021 [$-$0.025, $-$0.011] & $-$0.147 [$-$0.157, $-$0.139] & 0.006 \\
Strands 2B & XLM-EMO\textsubscript{in} & $-$0.046 [$-$0.052, $-$0.025] & +0.013 [$-$0.009, 0.034] & 1.000 & $-$0.010 [$-$0.013, 0.009] & +0.046 [0.030, 0.063] & 0.006 \\
Strands 2B & XLM-EMO\textsubscript{out} & $-$0.108 [$-$0.112, $-$0.089] & $-$0.122 [$-$0.136, $-$0.108] & 0.006 & $-$0.015 [$-$0.023, $-$0.004] & $-$0.059 [$-$0.071, $-$0.049] & 0.006 \\
\bottomrule
\end{tabular}
\end{adjustbox}
\caption{Paired group bootstrap of the differences in calibration (complementary; $\Delta = A - B$, a negative $\Delta$ means that $A$ is better calibrated): expected calibration error and multiclass Brier score as returned and after temperature scaling (Table~\ref{tab:calibration}), with the calibration temperature re-estimated in every resample. Same resamples as for macro-F1; values averaged over the test sets common to the two systems (eight for XLM-EMO). $p_{\text{Holm}}$: bootstrap $p$-value of the Brier difference, Holm-adjusted over the 28 tests of the table; the ECE has intervals only. The comparisons and the procedure were fixed before the intervals were computed, but after the point estimates were known (\statref{}).}
\label{tab:calibration-tests}
\end{table*}

%% file: tables/latency.tex
\begin{table*}[t]
\centering
\small
\setlength{\tabcolsep}{4pt}
\begin{adjustbox}{max width=\linewidth}
\begin{tabular}{lllrrrr}
\toprule
System & Measured on & GPU (variants) & Median & Mean & 95th pct. & Records \\
\midrule
Jev & API & -- & 220 & 225 & 273 & 50{,}532 \\
OpenAI Decisions & API & -- & 229 & 210 & 280 & 50{,}532 \\
GPT-6 Luna & API & -- & 1{,}107 & 1{,}479 & 3{,}223 & 50{,}532 \\
\midrule
Strands 2B & GPU, 1 request & A10G (1); L40S (3) & 65 & 67 & 72 & 50{,}532 \\
Qwen3.5-2B & GPU, 1 request & L4 (3); L40S (1) & 146 & 152 & 182 & 50{,}532 \\
\midrule
tabularisai\textsubscript{in/out} & GPU, batch 32 & T4 & 2.5 & 3.0 & 5.2 & 378{,}990 \\
XLM-EMO\textsubscript{in/out} & GPU, batch 32 & T4 & 3.3 & 3.5 & 5.3 & 205{,}824 \\
\midrule
\multicolumn{7}{l}{\emph{Complementary runs (Section~\ref{sec:complementary}), GPU, 1 request}} \\
Qwen3.5-2B, LS & GPU & L4 (4) & 91 & 92 & 106 & 50{,}532 \\
Qwen3.5-2B, English & GPU & L4 (4) & 143 & 151 & 175 & 50{,}532 \\
Qwen3.5-2B, LS, English & GPU & L4 (4) & 84 & 85 & 102 & 50{,}532 \\
Strands 2B, English & GPU & L4 (4) & 62 & 62 & 64 & 50{,}532 \\
Qwen3.5-4B, LS & GPU & L4 (1) & 178 & 189 & 236 & 12{,}633 \\
Qwen3.5-4B, LS & GPU & A10G (3) & 176 & 172 & 210 & 37{,}899 \\
Qwen3.5-4B, generative & GPU & A10G (4) & 209 & 218 & 255 & 50{,}532 \\
\bottomrule
\end{tabular}
\end{adjustbox}
\caption{Latency per text in milliseconds, over all stored records (zero-shot systems: 4 variants; fine-tuned encoders: all in-domain and out-of-domain models and seeds). API: measured on the client, network included; GPU: one request at a time, after a warm-up, with the GPU of each variant in parentheses; batch 32: batch time divided by 32. OpenAI Decisions has a bimodal distribution, with about 30\% of the requests between 90 and 170\,ms in every variant and test set, so its mean is below its median. The three groups are measured differently and are not comparable with each other (Section~\ref{sec:latency}); within the API group, concurrency was 4 requests for the decision APIs and 16 for GPT-6 Luna. The complementary runs use Italian instructions unless stated otherwise; LS: label scoring. With English instructions all four variants of the 2B models ran on the L4 GPU, which gives a comparison on the same GPU.}
\label{tab:latency}
\end{table*}